\documentclass[10pt,journal,compsoc]{IEEEtran}
\usepackage{epsfig}
\usepackage{graphicx}
\usepackage{amsmath}
\usepackage{amssymb}
\usepackage{multicol}
\usepackage{multirow}
\usepackage{enumitem}
\usepackage{nccmath}
\usepackage{pifont}
\usepackage{multirow}
\usepackage{array}
\usepackage{adjustbox}
\usepackage{enumitem}
\usepackage{caption}
\usepackage[table,xcdraw,dvipsnames]{xcolor}
\usepackage{booktabs}
\usepackage{ragged2e}

\newcolumntype{x}[1]{>{\centering\arraybackslash\hspace{0pt}}p{#1}}

\makeatletter
\def\thickhline{%
  \noalign{\ifnum0=`}\fi\hrule \@height \thickarrayrulewidth \futurelet
   \reserved@a\@xthickhline}
\def\@xthickhline{\ifx\reserved@a\thickhline
               \vskip\doublerulesep
               \vskip-\thickarrayrulewidth
             \fi
      \ifnum0=`{\fi}}
\makeatother

\newlength{\thickarrayrulewidth}
\definecolor{darkgreen}{rgb}{0.0, 0.2, 0.13}
\definecolor{darkspringgreen}{rgb}{0.09, 0.45, 0.27}

\usepackage[colorlinks=false,citecolor=black]{hyperref}

\usepackage{xspace}

\makeatletter
\DeclareRobustCommand\onedot{\futurelet\@let@token\@onedot}
\def\@onedot{\ifx\@let@token.\else.\null\fi\xspace}

\def\eg{\emph{e.g}\onedot} 
\def\ie{\emph{i.e}\onedot}

\def\etal{\emph{et al}\onedot}
\makeatother

\ifCLASSOPTIONcompsoc
  \usepackage[nocompress]{cite}
\else
  \usepackage{cite}
\fi
\ifCLASSINFOpdf
\else
\fi
\begin{document}
%
\title{DSGN++: Exploiting Visual-Spatial Relation for Stereo-based 3D Detectors}
%
%
%
%

\author{Yilun~Chen, \textit{Student Member, IEEE}, Shijia~Huang, \textit{Student Member, IEEE}, \\ Shu~Liu, \textit{Member, IEEE}, Bei~Yu, \textit{Member, IEEE}, Jiaya~Jia, \textit{Fellow, IEEE}

\IEEEcompsocitemizethanks{\IEEEcompsocthanksitem Y. Chen, S. Huang, B. Yu and J. Jia are with the Department
of Computer Science and Engineering, The Chinese University of Hong Kong, Hong Kong, China.\protect\\
E-mail: \texttt{\{ylchen,sjhuang,byu,leojia\}@cse.cuhk.edu.hk; liushuhust@gmail.com} }
\IEEEcompsocitemizethanks{\IEEEcompsocthanksitem J. Jia and S. Liu are with the SmartMore.}
}

\IEEEtitleabstractindextext{%
\begin{abstract} 
\justifying  Camera-based 3D object detectors are welcome due to their wider deployment and lower price than LiDAR sensors. We first revisit the prior stereo detector DSGN for its stereo volume construction ways for representing both 3D geometry and semantics. We polish the stereo modeling and propose the advanced version, DSGN++, aiming to enhance effective information flow throughout the 2D-to-3D pipeline in three main aspects. First, to effectively lift the 2D information to stereo volume, we propose depth-wise plane sweeping (DPS) that allows denser connections and extracts depth-guided features. Second, for grasping differently spaced features, we present a novel stereo volume -- Dual-view Stereo Volume (DSV) that integrates front-view and top-view features and reconstructs sub-voxel depth in the camera frustum. Third, as the foreground region becomes less dominant in 3D space, we propose a multi-modal data editing strategy -- Stereo-LiDAR Copy-Paste, which ensures cross-modal alignment and improves data efficiency. Without bells and whistles, extensive experiments in various modality setups on the popular KITTI benchmark show that our method consistently outperforms other camera-based 3D detectors for all categories. Code is available at \url{https://github.com/chenyilun95/DSGN2}.
\end{abstract}

}

\maketitle

\IEEEdisplaynontitleabstractindextext

%
\IEEEpeerreviewmaketitle

\IEEEraisesectionheading{\section{Introduction}\label{sec:introduction}}

\IEEEPARstart{C}{amera}-based 3D visual perception is a fundamental and challenging task in 3D computer vision, which serves as the essential component for autonomous driving and robotics. 

The main difficulty of camera-based 3D detectors lies in the fact that \textit{cameras provide front-view information but generally lack top-view cues or depth for accurate 3D object localization}. A common choice for camera-based 3D detectors is to leverage the successful 2D object detectors  \cite{rcnn, fastrcnn, fasterrcnn, ssd} and depth estimators \cite{dorn, psmnet, gcnet}. A series of approaches \cite{stereorcnn,qin2019monogrnet,MLF,zhang2021objects,ding2020learning,triangulation,disentangling,brazil2019m3d,fcos3d} design complicated strategies to predict 3D boxes with explicit projective geometry of keypoints or boxes. 

In contrast, the 2D-to-3D transformation converts the problem on 3D representation \cite{oftnet, pseudolidar, chen2020dsgn}, which sidesteps the dimensionality loss in solving 3D problems. Particularly, the problem of 3D detection can be solved innately by predicting objects over every 3D spatial location. For instance, Pseudo-LiDARs \cite{pseudolidar, accuratemono3d, pseudo++} generate explicit 3D representation followed by direct application of 3D detectors \cite{pointrcnn, second, VoxelNet}.  The explicit 3D form via, \eg, depth maps, occupancy grids, or pseudo point clouds, removes the uncertainty and decouples the tasks of depth estimation and object recognition. 

However, predicting depth from images is ill-posed. Thus, the generated depth cost volume depicts the uncertainty of voxel occupancy. To preserve the knowledge of depth uncertainty, implicit modeling of geometry-encoded feature volume becomes popular recently \cite{chen2020dsgn, plume, endtoendpseudolidar, CaDDN, guo2021liga}. Stereo geometry (or epipolar geometry) is encoded into the concrete 3D voxel grids and guides the following 3D prediction. In essence, feature transformation from 2D to 3D representation avoids the loss of geometric uncertainty, which is proved influential for following 3D prediction \cite{chen2020dsgn}. 

Therefore, valid information flow from 2D semantics to stereo volume determines the efficacy of the following 3D geometric representation for both geometric and semantic cues. However, current 3D modeling remains as an approximation of realistic 3D representation and poses three vital challenges for creating effective stereo feature volumes as follows:

(I) \textit{Direct 2D-to-3D information propagation along the ray constrains volumetric representation power}. In geometric modeling, \textit{plane sweeping} (PS) \cite{planesweep, deepstereo, mvsnet} is the dominant way to lift 2D information to 3D volume. Specifically, per-view features are directly propagated by tracing the ray in the volumetric space \cite{gcnet, psmnet} for matching the pixel differences. As 3D volume consumes one more order of magnitude of computation cost than 2D one, raw 2D features are required to be compressed to a small channel to reduce the amount of calculation. Accordingly, the 3D representation power is limited by the compressed 2D features. Our finding reveals that alleviating this bottleneck unleashes the power of volumetric representation for the following prediction tasks. 

(II) \textit{Perception of differently shaped objects}. By plane sweeping, we can produce two views of stereo volumes: Plane-sweep volume (PSV) in camera frustum and 3D-geometry volume (3DGV) in regular 3D space. However, real-world 3D objects are non-rigid and irregular-shaped. Some categories like \textsl{Pedestrian} occupy fewer voxels in the bird's eye view albeit being clearly visible in the front view.  In our study, these single-view stereo volumes show varied properties. Plane-sweep volume extracts more voxel features for front-view objects (such as \textsl{Pedestrian} and \textsl{Cyclist}) while 3D-geometry volume obtains same gradients for same object at different distances. 

(III) \textit{Biased modeling}. The proportion of the foregrounds is usually small in the aerial view for outdoor scenes, which curbs data efficiency. Additionally, imbalanced class distribution also gives the biased gradient flow towards frequent objects and suppresses the generalization ability of stereo modeling. These two difficulties restrict the model's capacity to generate unbiased estimation.

In this paper, we provide the following three solutions for addressing the above challenges in 2D-to-3D modeling. By polishing the overall stereo modeling, we present a simple yet effective stereo-based 3D detection framework DSGN++.

First, we present a generic operator for 2D-to-3D transformation -- \textit{depth-wise plane sweeping} (D-PS) to relieve the bottleneck of 2D-to-3D information propagation. With D-PS, the transformation allows the input of wider 2D features, that encodes depth-guided features within its expanded channels. And the generated volume yields \textit{continuously changing} features that slice the 2D features via the sliding window technique.
A key component called ``cyclic slicing'' is employed to realize \textit{local feature continuity} for nearby depth planes. Experiments demonstrate its notable improvement for both monocular and binocular camera-based 3D object detection.

Second, we provide a new form of stereo volumetric representation -- Dual-view Stereo Volume (DSV) to build more extensive connections to different views. We aggregate features of differently-shaped voxels from the front view (plane-sweep volume) and top view (3D-geometry volume). Notably, with a front-surface depth head, the final cost volume is generated by transforming volumetric representation to camera frustum space because we found that the geometric supervision of sub-voxel depth values in the front view provides stronger supervision than discretized voxel occupancy learning. 

Last, to overcome the limited foreground regions in 3D modeling and make unbiased predictions towards categories, we seek to apply the copy-paste strategy \cite{copypaste}. However, the requirement of precise cross-modal alignment restricts the freedom of data editing. To overcome the limitation, we propose \textit{Stereo-LiDAR Copy-Paste} (SLCP) that allows joint stereo and 3D data editing and meets the constraint of cross-modal projection. We validate this flexible data editing improves modeling efficiency and generalization ability to various categories.

Our total contribution is fourfold. 
\begin{itemize}[itemsep=2.mm, leftmargin=*]
	\item Without additional computation, we propose a novel volume construction way of \textit{depth-wise plane sweeping} (D-PS) to expand the capacity of information flow and extract depth-relevant 2D features. 
	\item We propose Dual-view Stereo Volume with the front-surface depth head to extract the features from two differently spaced stereo volumes and investigates its effectiveness over prior constructions.
	\item For the first time, the method augments multi-modal data pair by \textit{Stereo-LiDAR Copy-Paste} strategy that ensures the stereo alignments at the sub-pixel level and improves the data efficiency. We prove that the strategy greatly mitigates the class imbalanced problem.
	\item Without bells and whistles, our proposed DSGN++ achieves the \textbf{first} place for \textbf{all categories} among all camera-based approaches on the challenging KITTI benchmark \cite{kitti} on Nov 20, 2021, and even surpasses some LiDAR detectors in AP$_{3D}$, such as AVOD \cite{AVOD} for the first time. 
\end{itemize}

\section{Related Work}

\noindent \textbf{Stereo Matching and Multi-View Stereo.~~} With the development of neural networks in stereo matching, methods of \cite{gcnet, psmnet, ganet, groupwisestereo, pwcnet, anytimestereo} process the left and right images by a Siamese network and construct a 3D cost volume to compute the matching cost. Correlation-based cost volume is applied in recent work \cite{dispnet, hierarchical, segstereo, groupwisestereo,learndispconsistency,edgestereo}. Methods of \cite{gcnet, psmnet} form a concatenation-based cost volume and apply 3D convolution to regress disparity estimates. For multi-view scene reconstruction, prior work \cite{mvsnet, MVSMachine, r-mvset, point-mvsnet, surfacenet, deepmvs, casmvsnet, MVSMachine} even achieves fewer depth errors than RGB-D sensors, which shows great potential to be an alternative of expensive depth sensors. MVSMachine \cite{MVSMachine} proposes the differentiable projection and unprojection for better extracting 3D  to manipulate the volume construction from multi-view images. 

\noindent \textbf{LiDAR-based 3D Detection.~~} LiDAR sensors are very powerful to produce data for 3D detectors. The target of LiDAR-based detectors is to extract discriminative features from point clouds for 3D object recognition. There are generally two types of 3D representations, i.e., voxel-based representation \cite{VoxelNet,MV3D, pointpillar, fastpointrcnn} and point-based representation \cite{pointnet,pointnet++,pointrcnn,std,pointfusion}.  Albeit depth sensors (\eg, LiDAR sensors and RGB-D cameras) can retrieve accurate depth cues, they are generally more expensive and are with sparser sensing resolution than the common off-the-shelf RGB cameras. We prove that with a simple fusion strategy, our stereo modeling can further promote the performance of LiDAR-based 3D detectors.


\noindent \textbf{Camera-based 3D Detection.~~} In contrast to the high cost and sparse resolution of depth sensors, cameras are readily available and applied on a wide scale. The dense imaging resolutions ($>720$P) provide human-readable semantics that is easy to distinguish. The accessibility and dominance as the basic perception sensor in the real world make it attractive to perceive and understand 3D scenes. We classify methods into two types according to their intermediate 2D or 3D representation. The key difference is that 2D representation extracts features in the front view while 3D or bird's eye view (BEV) type extracts features in top views or 3D space. 

For 2D representation-based 3D detectors, an intuitive solution is to leverage a 2D object detector \cite{fasterrcnn, ssd}. Similar to 2D object detectors, prior work \cite{stereorcnn, qin2019monogrnet, disentangling, brazil2019m3d, monopair, shi2021geometry, shi2020distance, centernet, fcos3d, zhang2021objects} directly estimates 3D bounding boxes from camera images and relies on perspective modeling of the 2D projected object and its 3D objects. On the other hand, depth supervision via point clouds or depth maps is accessible during training. Explicit learning of depth cues improves the accuracy of 3D object detection \cite{ma2021delving}. Methods of \cite{messagepropagation, MLF, ding2020learning} also jointly aggregate the learned depth cues and semantic cues. 


Generally, due to the consistency with 3D spaces, the 3D form provides an elegant and effective representation with no complicated post-processing steps. 3DOP \cite{3dop,3dop-pami} generates point clouds by stereo and encodes the prior knowledge and depth in an energy function. Several methods \cite{pseudolidar, pseudo++, accuratemono3d} transform the depth map to Pseudo-LiDAR with point cloud followed by another independent network. Pseudo-LiDAR~\cite{pseudolidar, pseudo++, accuratemono3d, ocstereo} introduces the pseudo point clouds as the intermediate 3D representation followed by a LiDAR-based 3D detector. This pipeline yields much improvement over previous 2D representation-based approaches. E2E-PL~\cite{endtoendpseudolidar} further enables back-propagation to depth coordinates by introducing radial basis functions. We note that these methods are limited to explicit modeling of depth maps. They compress the abundant information from pixel-level feature projection and correspondence. Recent end-to-end pipelines \cite{oftnet, chen2020dsgn, plume, CaDDN, philion2020lift, div2020wstereo} utilize 3D feature volumes as intermediate representation. Recently DSGN \cite{chen2020dsgn} performs remarkably by implicitly encoding 3D geometry into neural networks. CDN \cite{div2020wstereo} further refine depth prediction near object boundarys via a Wasserstein distance-based loss. PLUME \cite{plume} designs the efficient 3D-BEV network to achieve proper trade-off between speed and accuracy. LIGA-stereo \cite{guo2021liga} further leverages the well-learned LiDAR-detector to transfer the knowledge to DSGN and demonstrates the effectiveness of cross-modal distillation \cite{crossmodaldistillation}.

\section{Our Approach}

\begin{figure*}
    \begin{center}
        \includegraphics[width=1.\linewidth]{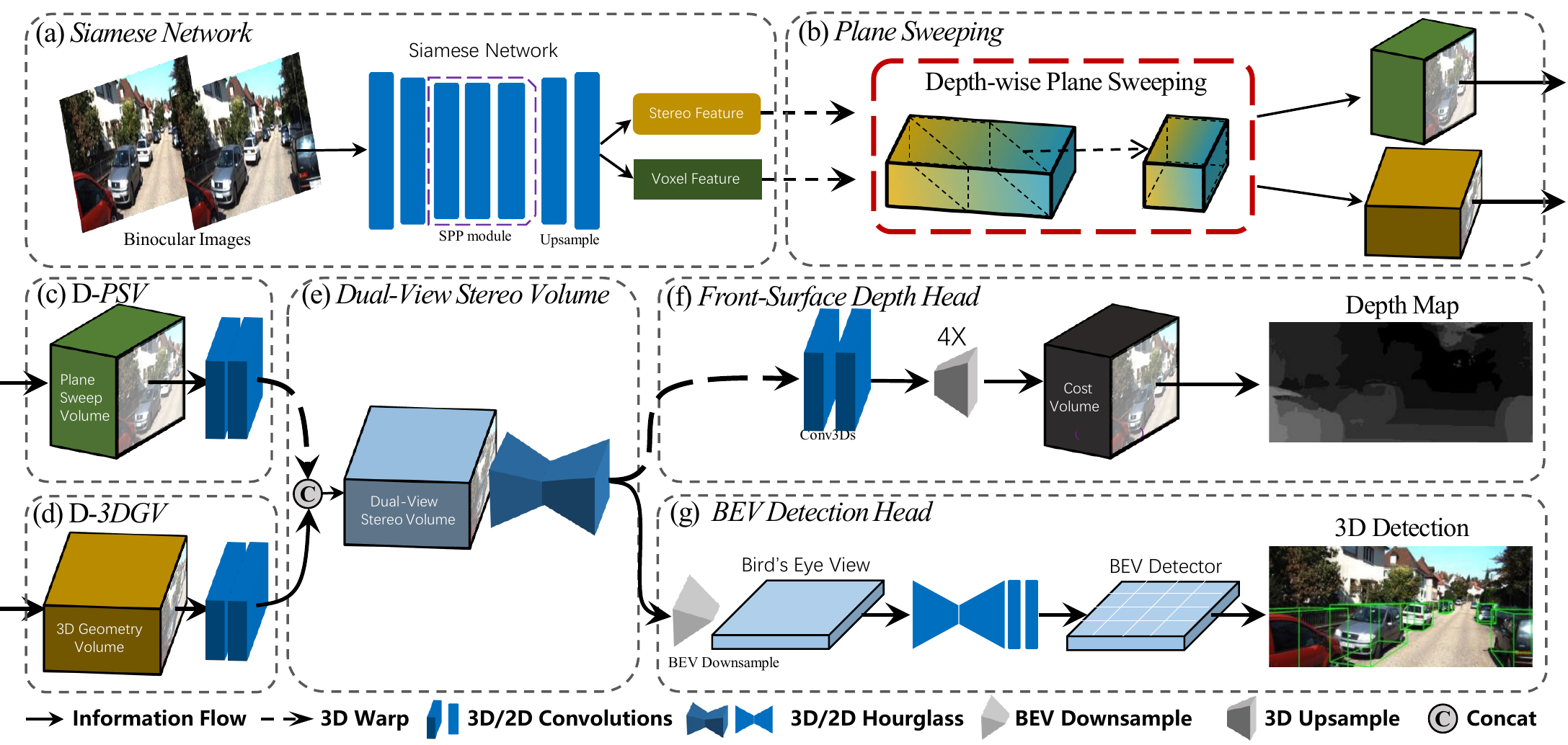}
    \end{center}
    \caption{\textbf{Overview of the proposed DSGN++ framework}. The whole framework consists of six components. (a) 2D image extraction network for extracting stereo features. (b) Volume construction process by Depth-wise Plane Sweeping. (c) 3D CNN for front-view and top-view feature extraction. (d) Dual-view flow integration followed by 3D CNN. (e) Front-surface depth head for supervising depth signals in the front-view. (f) 3D detection head that detects objects in bird's eye view.}
    \label{fig:pipeline}
	\vspace{-0.1in}
\end{figure*}

In Sec.~\ref{sec: background of volumes}, as a prerequisite, we revisit the stereo volume generation and transformation in DSGN \cite{chen2020dsgn} for encoding implicit cues of geometry and semantics.
We introduce our DSGN++ model (shown in Fig.~\ref{fig:pipeline}) to increase the capacity of stereo modeling in the following three aspects. 

First, we identify the network bottleneck that limits the quantity of information flow and introduce a generic operator -- depth-wise plane sweeping (D-PS) (Sec.~\ref{sec: depth-wise plane sweeping}) for allowing denser connections between 2D and stereo volumes.

Second, in Sec.~\ref{sec: dsgn++}, we compare the volume effectiveness for differently shaped objects between camera front-view and top-view. For aggregating more view-specific features, we introduce Dual-view Stereo Volume (DSV), which includes volumes integration and front-surface depth head. 

Finally, in Sec.~\ref{sec: stereo copy paste}, we introduce a multi-modal data augmentation strategy -- joint \textit{Stereo-LiDAR copy-paste} for increasing the positive ratios and balancing the category distribution in each training sample.

\subsection{Stereo Volumes Generation Revisit} \label{sec: background of volumes}
Given a binocular image pair ($I_L, I_R$), the objective is to detect and localize objects in 3D world space. 
To avoid information loss of depth uncertainty in explicit data structures such as point clouds, recent approaches \cite{chen2020dsgn, plume} create stereo volumes and encode the geometry cues into the 3D feature volume. With accurate depth cues, the 3D detection head can detect and regress 3D objects efficiently, especially for faraway objects. In this section, we briefly revisit several structures of stereo feature volumes -- plane-sweep volume in camera frustum space and 3D-geometry volume (or BEV volume \cite{plume}) in 3D regular space. 

For simplicity, we denote the voxel coordinate $\mathbf{u} = \left( u, v, d \right)$ in the camera frustum and the voxel coordinate $\mathbf{x} = \left( x, y, d \right)$ in pre-defined voxel space, where $d$ denotes the depth dimension. $proj: \mathbb{R}^3 \rightarrow \mathbb{R}^2$ represents the projection.

\noindent \textbf{Binocular Images to Plane-Sweep Volume.~~} The binocular features are generated by feeding a binocular image pair ($I_L, I_R$) into a Siamese network. In stereo matching \cite{gcnet, flownet, deepstereo} and MVS \cite{mvsnet, MVSMachine}, a set of evenly-spaced depth (or disparity) planes are generated towards the target view. By the classic sweeping planes towards the camera view, multiplane images (MPI) \cite{stereomagnification} are generated by gathering image features at each depth plane. The per-view mapping function can be formulated as
\begin{equation}
\begin{split}
    &PSV: \mathbb{R}^{H_I\times W_I\times C_I} \rightarrow \mathbb{R}^{{   \iffalse \color{red} \fi    H_I'\times W_I'}\times D_V\times C_V}, PSV(\mathbf{I}) = \mathbf{V}_\text{proj}  \\
    &\text{where}\ \  \mathbf{V}_\text{proj}\left(\mathbf{u},c\right) = \mathbf{I}\left(proj(\mathbf{u}), c\right). 
\end{split}
\end{equation}
where {   \iffalse \color{red} \fi     the size of ($H_I', W_I'$) is linearly related to ($H_I, W_I$) }. Voxels with coordinates $\{\mathbf{x} = (u,v,d) \}_{PSV}$ are uniformly spaced in the camera frustum with $C_V$ ($C_V= C_I$ channels. By comparing feature similarity of each voxel, the following neural network infers the underlying 3D geometry at the target view.


\noindent \textbf{Plane-Sweep Volume to 3D-Geometry Volume.~~} 
As the final objective is to detect 3D objects in 3D world space, one way to encode the scene in the 3D world is to transform PSV to 3DGV. Specifically, a detection area of size $(H_V, W_V, D_V)$ can be discretized into voxel gird. 3D-geometry volume is computed by reversing 3D projection from camera frustum space to 3D world space.


\noindent \textbf{Binocular Images to 3D-Geometry Volume.~~} 
Another way to construct 3D-geometry volume (or BEV volume \cite{plume}) is introduced by the operation of \textit{differentiable unprojection} \cite{MVSMachine}. The discrete grid $\left\{ (x,y,z) \right\}_{3DGV}$ obtains the projected 2D image features at $(u,v)$ by \textit{differentiable bilinear sampling}. Image features at different views are aggregated inside each voxel. 

\subsection{Depth-wise Plane Sweeping for 2D-to-3D Transformation}  \label{sec: depth-wise plane sweeping}

3D volume construction from the images is critical to represent a 3D scene for either monocular or multi-view settings. It facilitates a series of downstream 3D applications, \eg, stereo matching, novel view synthesis, and 3D object detection. Without loss of generality, for a predefined voxel grid of size $(H_V, W_V, D_V)$ in \textit{   \iffalse \color{red} \fi     arbitrarily voxelized space}, we retrieve multi-view features by the per-view projection of each voxel coordinate $\mathbf{p} = (x, y, z)$.
{   \iffalse \color{red} \fi     The general formulation from a camera view to a specific volume $\mathbf{V}$ is expressed as}
\begin{equation}
\begin{split}
    &\mathbf{V}:\ \ \mathbb{R}^{H_I\times W_I\times C_I} \rightarrow \mathbb{R}^{H_V\times W_V\times D_V\times C_V}, 
    \mathbf{V}(\mathbf{I}) = \mathbf{V}_{\text{proj}} \\
    &\quad\quad\quad\text{where}\ \  \mathbf{V}_\text{proj}(\mathbf{p}, c) = \mathbf{I}(proj(\mathbf{p}), c).
\end{split}
\end{equation}
where the 2D feature has the shape of $(H_I, W_I, C_I)$.

\noindent \textbf{2D-To-3D Modeling Bottleneck.~~} During the construction of stereo volume, an evident fact is that the extra dimension orthogonal to camera planes is generated. Normally, the feature grids are filled by replication of image features through viewing rays. The size of generated 4D tensor $\mathbf{V}$ is normally far larger than the source 2D feature tensor, expressed as
\begin{equation}
    H_V\times W_V \times D_V \times C_V \gg H_I\times W_I \times C_I. 
\end{equation}
In prior stereo networks \cite{gcnet, psmnet}, the number of depth planes ($D_V$) is usually large, \eg., $192\text{-D}$ and the feature resolution is maintained at least a quarter of the full resolution for matching at the pixel level. In other words, the representation power (a.k.a. {effective degree of freedom} \cite{bengio2011expressive, expressive}) of the constructed 3D volume is constrained by its narrow 2D feature (small feature channels).

An ideal way for building a denser connection is to expand the channel size $C_I$ of 2D features, which reduces the tensor dimension gap. However, expanding $C_V$ is not straightforward as expanding volume channels $C_V$ leads to more calculations in 3D. More, the transformation needs to maintain \textit{feature locality} for matching the left-right correspondence cost.

Accordingly, we introduce depth-wise (disparity-wise) plane sweeping (D-PS) to build denser connections between 2D feature maps and 3D feature volumes. Instead of compressing 2D channels to a small number, we preserve the number of channels $C_I$ at a relatively large number (\eg, 96) and slice the feature via a sliding window ($C_V$ channels) along the channel axis. The shift on the channel axis depends on pixel disparity (inverse depth) as that distant object recognition is sensitive to sub-pixel differences.  We empirically show in Sec.~\ref{sec: ablation study} that overcoming this challenge leads to considerable performance gain.

\begin{figure}
	\begin{center}
		\includegraphics[width=.48\textwidth]{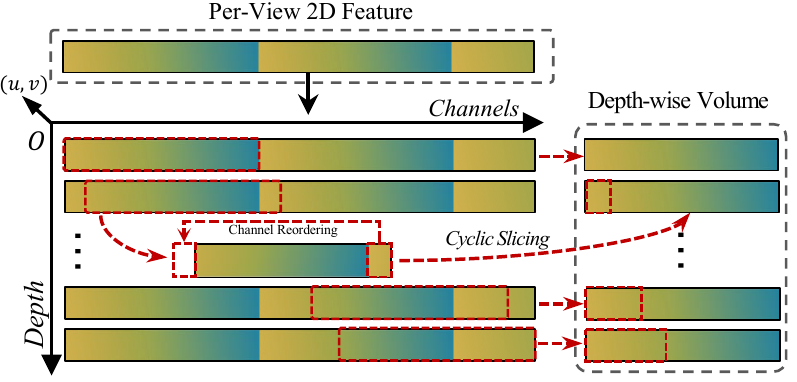}
	\end{center}
	\caption{\textbf{Depth-wise plane sweeping.} Assume the constructed volume is a 4D tensor in the space along $(x, y, \text{\textit{depth}}, \text{\textit{channel}})$ axes. We visualize the \textit{depth}-\textit{channel} plane of 3D featured volume where the graduated color indicates channel orders. Depth-wise volume is constructed by jointly sweeping the depth planes and slicing the features along the channel dimension. \textit{Cyclic slicing} reorders the channels to ensure channel consistency across nearby depth planes. }
	\label{fig:dps}
	\vspace{-0.1in}
\end{figure}

However, directly slicing the feature by shift produces unstable features as the order of feature channels is unchangeable. Therefore, we propose \textit{Cyclic Slicing} to ensure local feature similarity for adjacent objects, \ie, reordering the channels of the sliced features to maintain channel consistency.

\noindent \textbf{Cyclic Slicing.~~} Given a voxel coordinate $\textbf{x} = (x, y, d)$, with its image feature channel $C_I$, we obtain the feature slice with $C_V$ ($C_V \leq C_I$) channels as the voxel feature. As shown in Fig.~\ref{fig:dps}, we divide the $C_I$ channels into several parts. Each region contains $C_V$ channels except the last part. The sliced channels are an \textit{ordered} union of two channel intervals of
$$ \big [ \left \lceil disp / C_V \right \rceil \times C_V , disp + C_V \big) \cup \big [ disp, \left  \lceil disp / C_V \right \rceil \times C_V \big )$$
where $\left \lceil \  \right \rceil$ is the ceiling function. The reordering of the selected channels ensures feature continuity around nearby depth planes. For simplicity, we ignore cyclic slicing in Eqs. \eqref{equ: dpsv} and \eqref{equ: d3dgv}. The way to generate depth-wise plane-sweep volume (\textbf{D}-\textit{PSV}) is expressed as 
\begin{eqnarray}\scriptsize
&&\textbf{D}\text{-}PSV: \mathbb{R}^{H_I\times W_I\times C_I} \rightarrow \mathbb{R}^{{   \iffalse \color{red} \fi    H_I'\times W_I'}\times D_V\times C_V}\nonumber\\ &&\textbf{D}\text{-}PSV(\mathbf{I}) = \mathbf{V}_\text{proj} \label{equ: dpsv}
\end{eqnarray}
\text{where}
\begin{equation}\small
\mathbf{V}_\text{proj}(\mathbf{u}, c) = \mathbf{I}\left(proj(\mathbf{u}), \left\lfloor\frac{f_u \times baseline}{d}\right\rfloor^{\alpha} \mathbf{s} + c\right).\nonumber
\end{equation}
$\mathbf{s}$ is the ratio between 2D feature channels $C_I$ and the number of depth planes $D$. $\mathbf{u}$ denotes the coordinate $(u, v, d)$ in the camera frustum. $\alpha$ controls the smoothness of channel shifting rate. $f_u$ denotes horizontal focal length and $baseline$ denotes stereo camera baseline. 
Similarly, construction of depth-wise 3D-geometry volume (\textbf{D}-\textit{3DGV}) is formulated as
\begin{eqnarray}\small
&&\textbf{D}\text{-}\textit{3}DGV:\mathbb{R}^{H_I\times W_I\times C_I}\rightarrow\mathbb{R}^{H_V\times W_V\times D_V\times C_V}\nonumber\\ &&\textbf{D}\text{-}\textit{3}DGV(\mathbf{I}) = \mathbf{V}_\text{proj} \label{equ: d3dgv}
\end{eqnarray}
where
\begin{equation}\small
\mathbf{V}_\text{proj}(\mathbf{x}, c) = \mathbf{I}\left(proj(\mathbf{x}), \left\lfloor\frac{f_u \times baseline}{d}\right\rfloor^{\alpha} \mathbf{s} + c\right).\nonumber
\end{equation}
where $\mathbf{x}$ denotes the 3D coordinate $(x, y, z)$.

The computation complexity is exactly \textit{same} as classic plane sweeping despite the growth of memory usage (the expansion of 2D feature map size). Experiments show that the simple solution to reduce the bottleneck leads to a considerable performance boost without extra techniques. 

\subsection{Dual-view Stereo Volume for Building Effective 3D Representation}  \label{sec: dsgn++}

\begin{figure}
    \begin{center}
    \includegraphics[width=.916\linewidth]{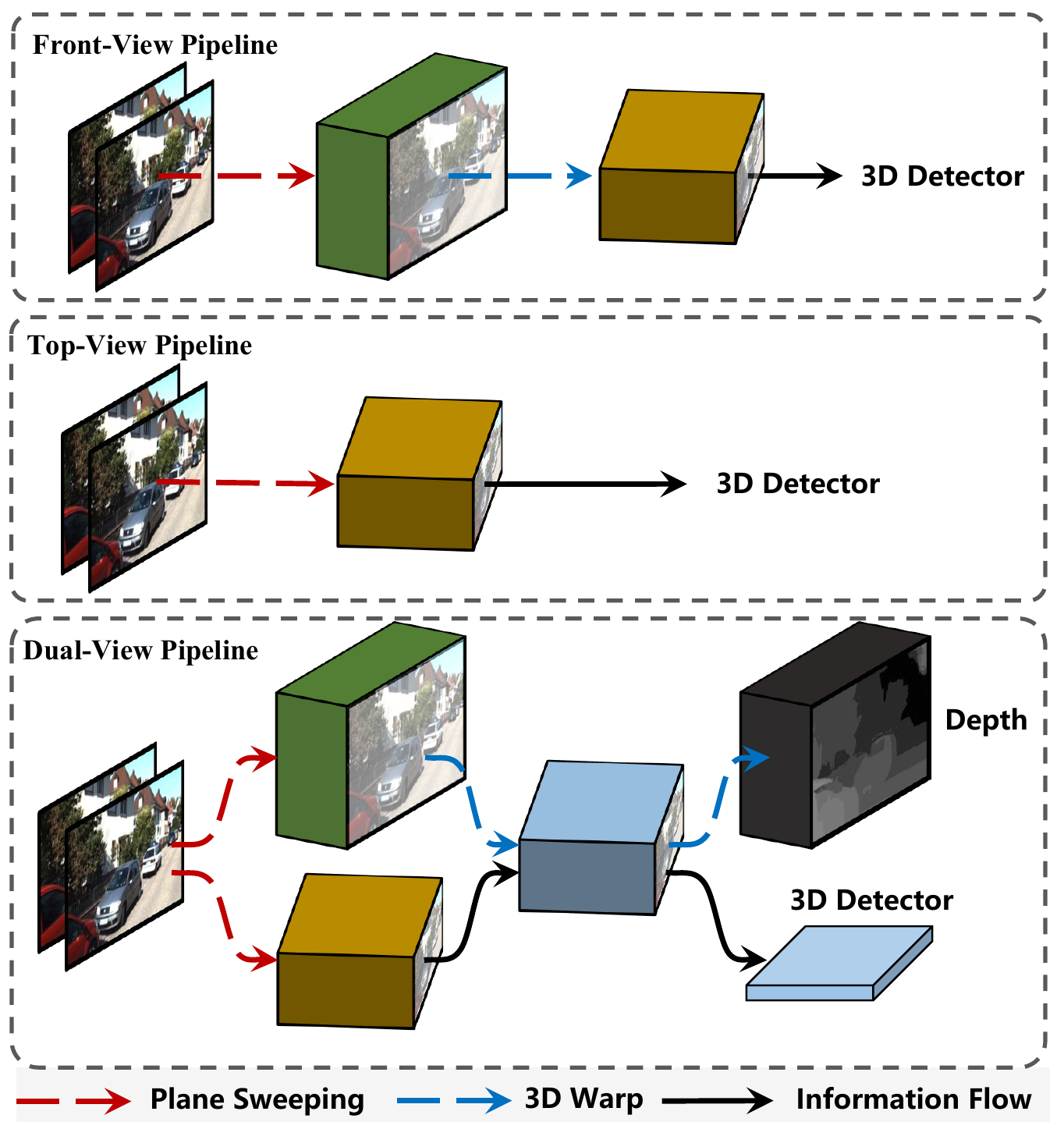}
    \end{center}
    \caption{\textbf{Comparison of stereo information flows.}  Prior stereo detectors adapt 2D features to plane-sweep volume (green cube) or 3D-geometry volume (golden cube). Differently, dual-view stereo volume (DSV) aggregates both spaced features in 3D space and enforces geometric learning in the front view that is fit for visual sensors. }
    \label{fig: different stereo volumes}
\end{figure}

In this section, we compare the information flows of two pipelines in recent works \cite{chen2020dsgn, plume, CaDDN, oftnet} as shown in Fig.~\ref{fig: different stereo volumes} (a, b) and analyze the difference between their voxel shapes. Further, to effectively represent depth and semantics, we introduce a new stereo volume -- Dual-view Stereo Volume (DSV), that is susceptible to both views. This volume construction contains two key steps: \textit{Volume Integration} and \textit{Front-Surface Depth Head}.

\begin{figure}
	\begin{center}
		\includegraphics[width=94mm]{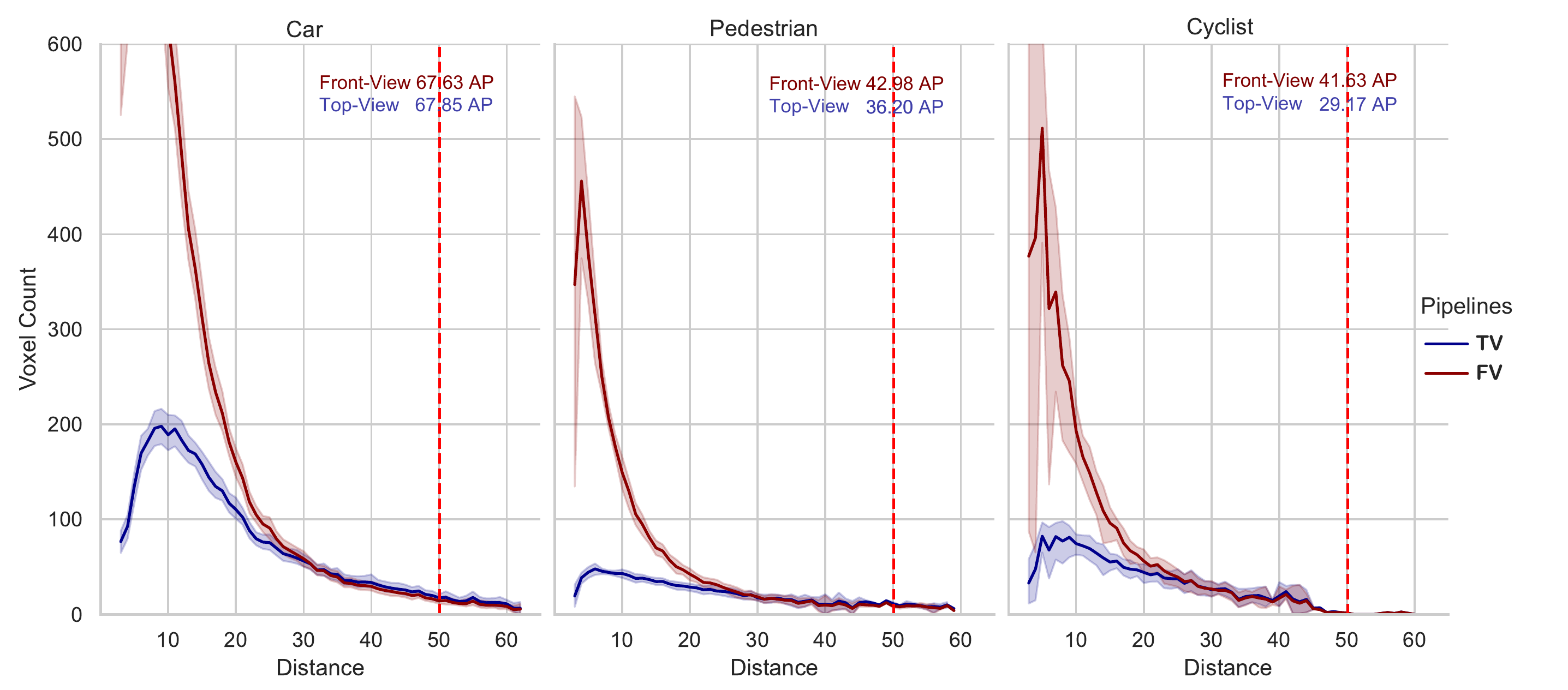}
	\end{center}
	\caption{\textbf{Comparison of average voxel occupancy per category within the plane-sweep and 3D-geometry volumes}. We set the maximum voxel numbers to $ 600$ for visualization of distant regions.}
	\label{fig:instace voxel density}
\end{figure}

\noindent \textbf{Front-view Representation vs. Top-view Representation.~~} 
In geometric learning, the front-view (FV) pipeline adopts plane-sweep volume for front-view depth learning in the camera frustum. Differently, the top-view (TV) pipeline constructs 3D structures within 3D-geometry volume (3DGV) in 3D regular space. The essential difference between the stereo volumes lies in their different shaped voxels or spaces, which directly leads to diverse receptive fields and voxel occupancy densities.  Hence, taking KITTI dataset~\cite{kitti} as an example, we visualize average voxel occupancy counts for all categories and their performance in Fig.~\ref{fig:instace voxel density}. Visually, nearby objects in PSV occupy much more voxels than faraway objects while the 3DGV curve is smoother. However, the average voxel occupancy counts for \textsl{Pedestrian} and \textsl{Cyclist} ($<20$m) are less than 100 voxels. The limited voxel occupancy impedes effective gradients towards the smaller objects, resulting in a performance reduction in the top-view pipeline. On the other hand, the distribution in PSV volume can deteriorate the learning of faraway objects.

\noindent \textbf{Stereo Volumes Integration.~~} After construction of $\textbf{D}\text{-}PSV$ and $\textbf{D}\text{-}3DGV$, we aggregate both information flows in our framework. Integration of both volumes allows the voxel to aggregate differently spaced 3D structure information and further expand the 2D-to-3D information flow. Specifically, we transform $\textbf{D}\text{-}PSV$ to 3D space and concatenate both volumes followed by a 3D Hourglass module \cite{psmnet}. The diversely spread voxel features are accessible within the combined feature volume. Experiments demonstrate the joint flows perform better than each independent volume under strong augmentation. 


\noindent \textbf{Front-Surface Depth Head.~~} \label{front-surface head}
Geometric learning determines the perception accuracy of distant scenes. DSGN \cite{chen2020dsgn} intermediately supervises the depth inside plane-sweep volume followed by the transformation to 3D-geometry volume that cannot benefit from the following computations. PLUME \cite{plume} adopts occupancy loss in the voxel grid that discretizes the depths, which is inferior to reason the geometry as shown in experiments (Sec. \ref{sec: ablation study}). 

To perceive accurate front-surface depths, the depth head on stereo volume in 3D space (\eg, DSV) is first transformed to the frustum space followed by front-view depth supervision. Meanwhile, the semantic supervision (e.g. 3D bounding boxes) jointly acts on the same feature volume. In detail, the transformation starts by building the 3D coordinate mapping from camera frustum space to 3D space. With the coordinate mapping, voxels at $(u, v, d)$ of the front-view volume obtain stereo volume features at $(x, y, z)$:
\begin{equation}
\begin{split}
\begin{pmatrix}
u\\v\\1
\end{pmatrix}d
&=
\begin{pmatrix}
f_u & 0 & c_u \\
0 & f_v & c_v \\
0 & 0 & 1
\end{pmatrix}
\begin{pmatrix}
x\\y\\z
\end{pmatrix}
\end{split},
\end{equation}
where $f_u, f_v$ are the horizontal and vertical focal lengths. {   \iffalse \color{red} \fi     We ignored the extrinsics for simplicity.} This generated front-view volume has the identical shape of PSV and is followed by a upsampling head network. The head network includes one hidden 3D convolution and a convolution that squeeze the channels to 1. The generated cost volume is then upsampled to original image size and supervised with depth loss \cite{chen2020dsgn, guo2021liga}. The construction enables mono-peak front-view depth that coincides with the real depth sensor data. Note that the transformation-based depth head is pluggable and can be inserted to 3D-geometry volume for both monocular and stereo settings.


\subsection{Stereo-LiDAR Copy-Paste for Improving Data Efficiency} \label{sec: stereo copy paste}

We illustrate the necessity to augment more and balanced foreground objects into the training scene as follows:

\noindent \textbf{Limited Foreground Area Ratio in Top View.~~} The 2D-to-3D transformation reduces the problem of front-view 3D detection to BEV detection. However, the foreground region ratio is also reduced in the bird's eye view. The imbalance decreases the magnitude of foreground gradients back to the 2D network, leading to biased model learning. 

\noindent \textbf{Imbalanced Class Distribution.~~} Long-tailed distribution commonly exists \cite{LVIS} in real scenes. For example, \textsl{pedestrian} and \textsl{cyclist} exists in less than 1/3 of full data, direct training of the imbalanced data could bias the gradient flow. 

\begin{figure}
	\begin{center}
		\includegraphics[width=85mm]{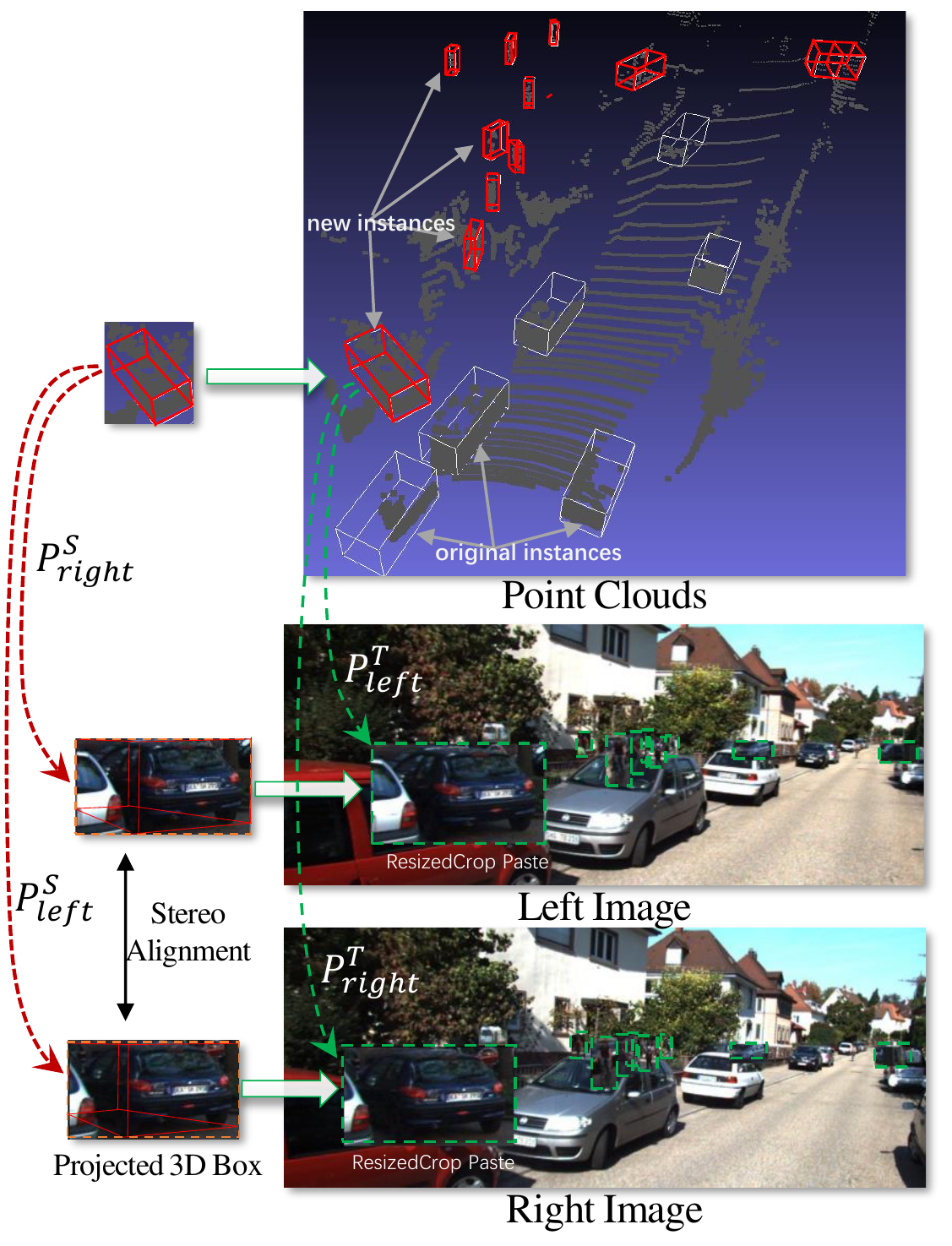}
	\end{center}
	\caption{\textbf{Joint \textit{Stereo-LiDAR copy-paste strategy}.~~} For binocular training samples, the source object patches are cropped with their calibrated projections $\left\{P^{S}_{left}, P^{S}_{right}\right \}$. The augmented scene uses projections $\left\{P^{T}_{left}, P^{T}_{right}\right \}$. Object patches are pasted bilinearly to binocular images. }
	\label{fig:stereo copy paste}
\end{figure}

Unlike point clouds, multi-modal data augmentation is constrained by the tight correspondence between image and point clouds. For the localization of 3D objects, the \textit{sub-pixel misalignment} affects the estimation of stereo disparity, leading to the large localization error for distant objects. Common copy-paste \cite{copypaste, cutpaste} randomly pastes object patches onto 2D images, which makes it hard to satisfy the projective constraint. Moreover, the segmentation mask is unavailable in binocular images and it is hard to guarantee the left-right alignment at pixel level without human annotations. 

To effectively augment the stereo data at the instance level, for the first time in literature, we propose a multi-modal data editing strategy -- \textit{Stereo-LiDAR copy-paste} (SLCP) that maintains precise cross-modal alignments at the \textit{sub-pixel} level. Specifically, we preserve the 3D location of the source objects and project the 3D boxes onto images with the target camera's internal parameters. As shown in Fig.~\ref{fig:stereo copy paste}, suppose we sample several objects from their source scene for a target training scene. The cropped point clouds within objects can be put into the training scene. For binocular training images, we compute the projected 2D bounding box $\mathbf{B}_S$ for each 3D object box $\mathbf{B}_{3D}$ by the source projection matrices $\{P^{S}_{left}, P^{S}_{right}\}$. By projecting the same 3D box by target projection $\left\{P^{T}_{left}, P^{T}_{right}\right \}$, we obtain the target bounding box $\mathbf{B}_T$. We crop and warp the source object image patches to the respective target boxes as $\mathbf{B}_S\rightarrow \mathbf{B}_T$. 
Note stereo alignment still holds under horizontal flipping. 
In this way, the alignment between LiDAR and both stereo images is guaranteed at the sub-pixel level. Also, to ensure \textit{uni-peak} depth, the overlapped 3D points, whose projections are within $\mathbf{B}_T$, are removed. 

For a training scene, we sample a sufficient and balanced number of objects per category (5 objects per category in our experiments) from other data and paste objects into the multi-modal data. Experiments demonstrate that our strategy effectively improves data efficiency and largely mitigates the imbalanced class distributions. 

{   \iffalse \color{red} \fi    \noindent \textbf{Comparison with geometry-preserved copy-paste}. Lian \etal \cite{lian2021geometry} introduce geometry-preserved way to paste segmented objects \cite{copypaste, cutpaste} for monocular 3D object detection. Instead of further position and size change under geometric constraints, our copy-paste simply maintains the 3D object locations for keeping both constraints of projection and stereo alignment. And our method also augments the corresponding point clouds within 3D boxes for LiDAR supervision.}

\begin{table*}[]
\begin{center}
\begin{tabular}{cllcccccccccc}
\toprule
\multirow{2}{*}{Sensor} & \multirow{2}{*}{Methods} & \multirow{2}{*}{Source} & {  \iffalse \color{red} \fi    L Sup.} & \multicolumn{3}{c}{\textsl{Car} \textbf{ AP$_{3D}$}} & \multicolumn{3}{c}{\textsl{Car}   AP$_{BEV}$} & \multicolumn{3}{c}{\textsl{Car}   AP$_{2D}$} \\ \cmidrule(lr){5-7} \cmidrule(lr){8-10} \cmidrule(lr){11-13}
 & &  &  & Easy & \textbf{Mod.} & Hard & Easy & \textbf{Mod.} & Hard & Easy & \textbf{Mod.} & Hard \\ \midrule
\multirow{5}{*}{LiDAR} & SECOND \cite{second} & Sensors2018 &  & 83.34 & 72.55 & 65.82 & 89.39 & 83.77 & 78.59 & -- & -- & -- \\
 & Point R-CNN \cite{pointrcnn} & CVPR2018 & & 86.96 & 75.64 & 70.70 & 92.13 & 87.39 & 82.72 & 94.00 & 91.90 & 88.17 \\
 & MV3D \cite{MV3D} & CVPR2017 &  & 74.97 & 63.63 & 54.00 & 86.62 & 78.93 & 69.80 & -- & -- & -- \\
 & AVOD \cite{AVOD} & IROS2018 &  & 76.39 & 66.47 & 60.23 & 89.75 & 84.95 & 78.32 & 95.17 & 89.88 & 82.83 \\
 & PL++: P-RCNN +SL* \cite{pseudo++} & ICLR2020 &  & 68.38 & 54.88 & 49.16 & 84.61 & 73.80 & 65.59 & 94.95 & 85.15 & 77.78 \\ \midrule
\multirow{14}{*}{Stereo} & TLNet~\cite{triangulation} & CVPR2019 & & 7.64 & 4.37 & 3.74 & 13.71 & 7.69 & 6.73 & 76.92 & 63.53 & 54.58 \\
 & Stereo-RCNN \cite{stereorcnn} & CVPR2019 & & 47.58 & 30.23 & 23.72 & 61.92 & 41.31 & 33.42 & 93.98 & 85.98 & 71.25 \\
 & PL: AVOD \cite{pseudolidar} & CVPR2019 &   \checkmark & 54.53 & 34.05 & 28.25 & 67.30 & 45.00 & 38.40 & 85.40 & 67.79 & 58.50 \\
 & ZoomNet \cite{zoomnet} & AAAI2020 &   \checkmark & 55.98 & 38.64 & 30.97 & 72.94 & 54.91 & 44.14 & 94.22 & 83.92 & 69.00 \\
 & PL++: P-RCNN \cite{pseudo++} & ICLR2020 &   \checkmark & 61.11 & 42.43 & 36.99 & 78.31 & 58.01 & 51.25 & 94.46 & 82.90 & 75.45 \\
 & OC-Stereo \cite{ocstereo} & ICRA2020 &   \checkmark & 55.15 & 37.60 & 30.25 & 68.89 & 51.47 & 42.97 & 87.39 & 74.60 & 62.56 \\
 & Disp R-CNN \cite{disprcnn} & TPAMI2021 &   \checkmark & 68.21 & 45.78 & 37.73 & 79.76 & 58.62 & 47.73 & 93.45 & 82.64 & 70.45 \\
 & DSGN \cite{chen2020dsgn}  & CVPR2020 &   \checkmark & 73.50 & 52.18 & 45.14 & 82.90 & 65.05 & 56.60 & 95.53 & 86.43 & 78.75 \\
 & CDN (DSGN) \cite{div2020wstereo} & NeurIPS2020 &   \checkmark & 74.52 & 54.22 & 46.36 & 83.32 & 66.24 & 57.65 & 95.85 & 87.19 & 79.43 \\
 & CG-stereo \cite{cgstereo} & IROS2020 &   \checkmark & 74.39 & 53.58 & 46.50 & 83.32 & 66.44 & 58.95 & 96.31 & 90.38 & 82.80 \\
 & YoLoStereo3D \cite{yolostereo3d} & AAAI2021 &   \checkmark & 65.68 & 41.25 & 30.42 & 76.10 & 50.28 & 36.86 & 94.81 & 82.15 & 62.17 \\
 & PLUME-Middle \cite{plume} & ICRA2021 &   \checkmark & -- & -- & -- & 83.0 & 66.3 & 56.7 & -- & -- & -- \\
 & LIGA \cite{guo2021liga}  & ICCV2021 &   \checkmark & 81.39 & 64.66 & 57.22 & 88.15 & 76.78 & 67.40 & 96.43 & 93.82 & 86.19 \\
 & DSGN++ (Ours) & -- &   \checkmark & \textbf{83.21} & \textbf{67.37} & \textbf{59.91} & \textbf{88.55} & \textbf{78.94} & \textbf{69.74} & \textbf{98.08} & \textbf{95.70} & \textbf{88.27} \\ \bottomrule
\end{tabular}
\end{center}
		\caption{\textbf{Performance comparison on the official KITTI \textit{test} server (\textsl{Car}).} * means refining the pseudo point clouds by additional 4-beam LiDAR. Best results are highlighted in \textbf{bold}. {  \iffalse \color{red} \fi    LiDAR supervision (L Sup.) represents whether to apply LiDAR depth supervision.}}
		\label{tab: KITTI complete test results}
\end{table*}

\section{Experiments}

In this section, we conduct extensive experiments to validate the effectiveness of our proposed framework with the settings of various modalities.
In Sec.~\ref{sec: experimental setup}, we briefly illustrate the datasets and experimental setups for 3D object detection. The quantitative and qualitative results are provided in Sec.~\ref{sec: main results} and Sec.~\ref{sec: visualization results}, respectively. We further illustrate the usefulness of each component in the ablation study (Sec.~\ref{sec: ablation study}). Lastly, the efficiency of our approach is discussed in Sec.~\ref{sec: speed}.

\subsection{Experimental Setup}\label{sec: experimental setup}
\noindent \textbf{Datasets.~~}
We evaluate methods on the popular KITTI 3D object detection dataset \cite{kitti}, which is a union of $7,481$ stereo image-pairs and point clouds for training and $7,518$ for testing. The training data has annotations for \textsl{Car}, \textsl{Pedestrian} and \textsl{Cyclist}. The ground-truth depth maps are generated from point clouds following \cite{pseudolidar,pseudo++,chen2020dsgn}. Following the protocol in \cite{MV3D, chen2020dsgn}, the training data is divided into a training set (3,712 images) and a validation set (3,769 images). As KITTI leaderboard limits the access to submission to the server for evaluating the test set, the ablation studies are conducted on the KITTI \textit{train}-\textit{val} split.  

\noindent \textbf{Evaluation Metric.~~}\label{evaluation metric}
KITTI divides the evaluation metric into three regimes (Easy, Moderate, and Hard) according to their recognition difficulty, which considers object occlusion/truncation and the size of an object in the 2D image. The AP evaluations for 2D, BEV and 3D have diverse IoU criteria per class , \ie, IoU $\geq$ 0.7 for \textsl{Car}, IoU $\geq$ 0.5 for \textsl{Pedestrian} and \textsl{Cyclist}. All experiments and ablation studies adopt $AP|_{R40}$ by default as the KITTI benchmark altered AP calculation that utilizes 40 recall positions ($AP|_{R40}$) instead of the earlier 11 recall positions ($AP|_{R11}$).  

\noindent \textbf{Experimental Setups.~~}\label{section:implementation} 
Our models are trained with the respectively best-performing parameters on four NVIDIA V100 GPUs, each GPU holding one pair of stereo images of size $384\times 1248$. We apply \textit{ADAM} \cite{adam} optimizer with initial learning rate 0.001. Data augmentation strategy \cite{chen2020dsgn} includes horizontal flipping and \textit{Stereo-LiDAR copy-paste}. All models are trained for 60 epochs and the learning rate is decreased by 10 at the 50-th epoch.

\noindent \textbf{Baseline Methods.~~}
We build our framework based on the official author-released codes -- DSGN \cite{chen2020dsgn} and LIGA \cite{guo2021liga}). LIGA {  \iffalse \color{red} \fi    reproduced} and improved DSGN in the code framework of OpenPCDet \cite{openpcdet} with several technical modifications for a stronger baseline. {  \iffalse \color{red} \fi    More implementation details are referred to the paper. We discard the strong cross-modal distillation technique throughout our experiments.} Based on the LIGA{  \iffalse \color{red} \fi   's reproduced} DSGN, we make several modifications for efficient implementation and adopt it (called L-DSGN) as the baseline approach unless otherwise specified. Particularly, we set the kernel size to $1\times 1\times 1$ for all the first 3D convolution after stereo volumes and move the 3D hourglass network after Dual-view Stereo Volume as shown in Fig. \ref{fig:pipeline}. Also, as depth-wise plane sweeping supports wider feature inputs, L-DSGN aggregates the multi-scale features by concatenation for binocular feature extraction instead of feature addition. The last 2D convolutional layer uses the number of filters $C_I$ according to the type of plane sweeping. During testing, the 2D detection head and depth prediction head are dropped. Synchronized batch normalization is applied throughout the network.

\noindent \textbf{Implementation of Stereo Volumes.~~} PSV is pre-defined with shape $(W_{I}/4, H_{I}/4, D_{I}/4, 64)$, where the image size is $(W_{I}=1248, H_{I}=384)$. Both left and right image features have 32 channels. The number of depth $D_{I}$ is set to $192$ (DSGN) and $288$ (L-DSGN). Extra 3D convolutions are applied to squeeze the channel dimension to $32$-D. 3DGV contains a 3D voxel occupancy grid of size $(W_{V}=300, H_{V}=20, D_{V}=288)$ along the respective directions in KITTI camera's view with each voxel of size $(0.2, 0.2, 0.2)$ (meters). Extra 3D convolutions with $32$ filters for compressing the features when generating 3DGV ($H_V\times W_V\times D_V\times 64$) directly from binocular images. Both stereo volumes adapt the 2D semantic features by plane-sweeping. We set the shifting ratio $\alpha$ of Depth-wise plane sweeping to 0.1 and its input channels $C=96$ by default.

\subsection{Quantitative Results}\label{sec: main results}
\subsubsection{Official Results on the KITTI \textit{test} benchmark~~} 
We report experimental results with comparison on the KITTI \textit{test} set as shown in TABLE \ref{tab: KITTI complete test results} and TABLE \ref{tab: KITTI complete test results ped cyc}. Without cross-modal distillation \cite{crossmodaldistillation, guo2021liga}, the simple solution DSGN++ outperforms {\it all} other stereo-based approaches over {\it all} difficulties and evaluation metrics (\textsl{Car}: +\textbf{2.71} AP$_{3D}$, +\textbf{2.16} AP$_{BEV}$, and +\textbf{1.88} AP$_{2D}$ in \textit{moderate} difficulty regime). In terms of AP$_{2D}$, our approaches achieve the impressively high \textbf{95.70} AP, surpassing all the strong 3D object detectors. 

As detecting smaller and non-rigid 3D objects pose the greater challenge for the regime of camera-based 3D detectors, only several approaches report the results for \textsl{Pedestrian} and \textsl{Cyclist}. To verify the generalization ability of our approach, we provide the results of \textsl{Pedestrian} and \textsl{Cyclist} in TABLE \ref{tab: KITTI complete test results ped cyc}. Our approach achieves noticeable improvements over prior methods (\textsl{Cyc.}: +\textbf{7.04} AP$_{3D}$, \textsl{Ped.}: +\textbf{2.74} AP$_{3D}$). 

Compared with LiDAR-based approaches, in terms of AP$_{3D}$, our method even completely beats some LiDAR detectors including AVOD for {\it all} categories (for the \textit{first} time in literature). Concretely, DSGN++ exceeds AVOD by 5.82 AP$_{2D}$ in the front view while scoring $6.01$ AP$_{BEV}$ lower than AVOD in the bird's eye view. This comparison indicates that the performance between camera-based approaches mainly lies in the foreground depth estimation error.

\begin{table*}[]
\begin{center}
\begin{tabular}{cllcccccccccccc}
\toprule
\multirow{2}{*}{Sensor} & \multirow{2}{*}{Methods} & \multirow{2}{*}{Source} & \multicolumn{3}{c}{\textsl{Ped.}   AP$_{3D}$} & \multicolumn{3}{c}{\textsl{Ped.}   AP$_{BEV}$} & \multicolumn{3}{c}{\textsl{Cyc.}   AP$_{3D}$} & \multicolumn{3}{c}{\textsl{Cyc.}   AP$_{BEV}$} \\ \cmidrule(lr){4-6} \cmidrule(lr){7-9} \cmidrule(lr){10-12} \cmidrule(lr){13-15}
 &  &  & Easy & \textbf{Mod} & Hard & Easy & \textbf{Mod} & Hard & Easy & \textbf{Mod} & Hard & Easy & \textbf{Mod} & Hard \\ \midrule
\multirow{2}{*}{LiDAR} & Point R-CNN   \cite{pointrcnn} & CVPR2018 & 47.98 & 39.37 & 36.01 & 54.77 & 46.13 & 42.84 & 74.96 & 58.82 & 52.53 & 82.56 & 67.24 & 60.28 \\
 & AVOD   \cite{AVOD} & IROS2018 & 36.10 & 27.86 & 25.76 & 42.58 & 33.57 & 30.14 & 57.19 & 42.08 & 38.29 & 64.11 & 48.15 & 42.37 \\ \midrule
\multirow{7}{*}{Stereo} & OC-Stereo   \cite{ocstereo} & ICRA2020 & 24.48 & 17.58 & 15.60 & 29.79 & 20.80 & 18.62 & 29.40 & 16.63 & 14.72 & 32.47 & 19.23 & 17.11 \\
 & Disp R-CNN   \cite{disprcnn} & TPAMI2021 & 37.12 & 25.80 & 22.04 & 40.21 & 28.34 & 24.46 & 40.05 & 24.40 & 21.12 & 44.19 & 27.04 & 23.58 \\
 & DSGN   \cite{chen2020dsgn} & CVPR2020 & 20.53 & 15.55 & 14.15 & 26.61 & 20.75 & 18.86 & 27.76 & 18.17 & 16.21 & 31.23 & 21.04 & 18.93 \\
 & CG-stereo   \cite{cgstereo} & IROS2020 & 33.22 & 24.31 & 20.95 & 39.24 & 29.56 & 25.87 & 47.40 & 30.89 & 27.73 & 55.33 & 36.25 & 32.17 \\
 & YoLoStereo3D   \cite{yolostereo3d} & AAAI2021 & 28.49 & 19.75 & 16.48 & 31.01 & 20.76 & 18.41 & -- & -- & -- & -- & -- & -- \\
 & LIGA   \cite{guo2021liga}  & ICCV2021 & 40.46 & 30.00 & 27.07 & 44.71 & 34.13 & 30.42 & 54.44 & 36.86 & 32.06 & 58.95 & 40.60 & 35.27 \\
 & DSGN++ (Ours) & -- & \textbf{43.05} & \textbf{32.74} & \textbf{29.54} & \textbf{50.26} & \textbf{38.92} & \textbf{35.12} & \textbf{62.82} & \textbf{43.90} & \textbf{39.21} & \textbf{68.29} & \textbf{49.37} & \textbf{43.79} \\ \bottomrule
\end{tabular}
\end{center}
\caption{\textbf{Performance comparison on the official KITTI \textit{test} server (\textsl{Pedestrian} and \textsl{Cyclist}).}  Best results are highlighted in \textbf{bold}. }
		\label{tab: KITTI complete test results ped cyc}
\end{table*}

\begin{table*}[bpt]\footnotesize
	\begin{center}
\begin{tabular}{cccccccccc}
\toprule
\multicolumn{1}{c}{} & \multicolumn{3}{c}{\textsl{Car}} & \multicolumn{3}{c}{\textsl{Pedestrian}} & \multicolumn{3}{c}{\textsl{Cyclist}} \\ \cmidrule(lr){2-4} \cmidrule(lr){5-7} \cmidrule(lr){8-10}
\multicolumn{1}{c}{\multirow{-2}{*}{Methods}} & \textbf{AP$_{3D}$} & AP$_{BEV}$ & \multicolumn{1}{c}{AP$_{2D}$} & \textbf{AP$_{3D}$} & AP$_{BEV}$ & \multicolumn{1}{c}{AP$_{2D}$} & \textbf{AP$_{3D}$} & AP$_{BEV}$ & AP$_{2D}$ \\ \midrule
\multicolumn{10}{l}{\textit{LiDAR Sensor}} \\ \midrule
\multicolumn{1}{c}{SECOND   \cite{second, openpcdet}} & 78.62 & 87.93 & \multicolumn{1}{c}{89.90} & 52.98 & 56.66 & \multicolumn{1}{c}{66.33} & 67.15 & 70.70 & 77.09 \\ 
\multicolumn{1}{c}{PV-RCNN   \cite{pvrcnn}} & 84.43 & 94.03 & 89.44 & 54.89 & 58.14 & 65.37 & 71.52 & 75.31 & 83.04  \\ \midrule
\multicolumn{10}{l}{\textit{Stereo Camera Sensor}} \\ \midrule
\multicolumn{1}{c}{DSGN   \cite{chen2020dsgn} $^\dag$} & 56.09 & 65.24 & \multicolumn{1}{c}{85.03} & 35.39 & 42.58 & \multicolumn{1}{c}{55.22} & 25.37 & 27.43 & 35.3 \\
\multicolumn{1}{c}{ours, DSGN++ on DSGN $^\dag$} & 61.62 & 70.61 & \multicolumn{1}{c}{89.47} & \textbf{44.17} & 48.51 & \multicolumn{1}{c}{62.35} & 36.04 & 39.05 & 43.56 \\
\multicolumn{1}{c}{{\iffalse \cellcolor{lightgray!40} \fi \textit{Improvement}}} & {\iffalse \cellcolor{lightgray!40} \fi \textit{+5.53}} & {\iffalse \cellcolor{lightgray!40} \fi \textit{+5.37}} & \multicolumn{1}{c}{{\iffalse \cellcolor{lightgray!40} \fi \textit{+4.44}}} & {\iffalse \cellcolor{lightgray!40} \fi \textit{+8.78}} & {\iffalse \cellcolor{lightgray!40} \fi \textit{+5.93}} & \multicolumn{1}{c}{{\iffalse \cellcolor{lightgray!40} \fi \textit{+7.13}}} & {\iffalse \cellcolor{lightgray!40} \fi \textit{+10.67}} & {\iffalse \cellcolor{lightgray!40} \fi \textit{+11.62}} & {\iffalse \cellcolor{lightgray!40} \fi \textit{+8.26}} \\ \midrule
\multicolumn{1}{c}{L-DSGN } & 63.58 & 73.53 & \multicolumn{1}{c}{93.59} & 33.12 & 40.50 & \multicolumn{1}{c}{59.16} & 28.09 & 29.58 & 36.95 \\
\multicolumn{1}{c}{ours, DSGN++ on L-DSGN} & \textbf{69.12} & \textbf{78.93} & \multicolumn{1}{c}{\textbf{95.85}} & 42.44 & \textbf{50.06} & \multicolumn{1}{c}{\textbf{68.92}} & \textbf{42.48} & \textbf{45.77} & \textbf{53.81} \\
\multicolumn{1}{c}{{\iffalse \cellcolor{lightgray!40} \fi \textit{Improvement}}} & {\iffalse \cellcolor{lightgray!40} \fi \textit{+5.54}} & {\iffalse \cellcolor{lightgray!40} \fi \textit{+5.40}} & \multicolumn{1}{c}{{\iffalse \cellcolor{lightgray!40} \fi \textit{+2.26}}} & {\iffalse \cellcolor{lightgray!40} \fi \textit{+8.32}} & {\iffalse \cellcolor{lightgray!40} \fi \textit{+8.56}} & \multicolumn{1}{c}{{\iffalse \cellcolor{lightgray!40} \fi \textit{+9.76}}} & {\iffalse \cellcolor{lightgray!40} \fi \textit{+14.39}} & {\iffalse \cellcolor{lightgray!40} \fi \textit{+16.19}} & {\iffalse \cellcolor{lightgray!40} \fi \textit{+16.86}} \\ \midrule
\multicolumn{1}{c}{  \iffalse \color{red} \fi    L-DSGN w/o LiDAR sup.} &  55.22 &  65.36 &  \multicolumn{1}{c}{  \iffalse \color{red} \fi    90.11} &  24.74 &  31.90 & \multicolumn{1}{c}{  \iffalse \color{red} \fi    49.49} &   21.67 &   23.04 &   34.27 \\
 \multicolumn{1}{c}{\scalebox{0.9}[1]{  \iffalse \color{red} \fi    ours, DSGN++ on L-DSGN w/o L Sup.}} & {  \iffalse \color{red} \fi    66.08} &   {75.92} &  \multicolumn{1}{c}{  \iffalse \color{red} \fi    95.52} &   31.49 &   39.30 & \multicolumn{1}{c}{ {  \iffalse \color{red} \fi    63.95}} &   {38.82} &   {40.54} &   {54.18} \\
\multicolumn{1}{c}{{\iffalse \cellcolor{lightgray!40} \fi \textit{  \iffalse \color{red} \fi    Improvement}}} &{\iffalse \cellcolor{lightgray!40} \fi \textit{  \iffalse \color{red} \fi    +10.86}} & {\iffalse \cellcolor{lightgray!40} \fi \textit{  \iffalse \color{red} \fi   +10.56}} & \multicolumn{1}{c}{{\iffalse \cellcolor{lightgray!40} \fi \textit{  \iffalse \color{red} \fi   +5.41}}} & {\iffalse \cellcolor{lightgray!40} \fi \textit{  \iffalse \color{red} \fi   +6.75}} & {\iffalse \cellcolor{lightgray!40} \fi \textit{  \iffalse \color{red} \fi   +7.40}} & \multicolumn{1}{c}{{\iffalse \cellcolor{lightgray!40} \fi \textit{  \iffalse \color{red} \fi   +14.46}}} & {\iffalse \cellcolor{lightgray!40} \fi \textit{  \iffalse \color{red} \fi   +17.15}} & {\iffalse \cellcolor{lightgray!40} \fi \textit{  \iffalse \color{red} \fi   +17.50}} & {\iffalse \cellcolor{lightgray!40} \fi \textit{  \iffalse \color{red} \fi   +19.91}} \\ \midrule
\multicolumn{10}{l}{\textit{Multi-Modal Sensors} } \\ \midrule
\multicolumn{1}{c}{4-LiDAR SECOND4x} & 23.82 & 30.85 & \multicolumn{1}{c}{32.27} & 16.70 & 21.79 & \multicolumn{1}{c}{24.85} & 12.18 & 13.45 & 14.05 \\
\multicolumn{1}{c}{Fusion with DSGN++} & 67.41 & 76.30 & \multicolumn{1}{c}{95.17} & 40.85 & 49.51 & \multicolumn{1}{c}{62.65} & 32.31 & 33.69 & 46.17 \\
\multicolumn{1}{c}{{\iffalse \cellcolor{lightgray!40} \fi \textit{Improvement}}} & {\iffalse \cellcolor{lightgray!40} \fi \textit{+43.59}} & {\iffalse \cellcolor{lightgray!40} \fi \textit{+45.45}} & \multicolumn{1}{c}{{\iffalse \cellcolor{lightgray!40} \fi \textit{+62.90}}} & {\iffalse \cellcolor{lightgray!40} \fi \textit{+24.15}} & {\iffalse \cellcolor{lightgray!40} \fi \textit{+27.72}} & \multicolumn{1}{c}{{\iffalse \cellcolor{lightgray!40} \fi \textit{+37.80}}} & {\iffalse \cellcolor{lightgray!40} \fi \textit{+20.13}} & {\iffalse \cellcolor{lightgray!40} \fi \textit{+20.24}} & {\iffalse \cellcolor{lightgray!40} \fi \textit{+32.12}} \\ \midrule
\multicolumn{1}{c}{8-LiDAR SECOND4x} & 49.00 & 66.83 & \multicolumn{1}{c}{66.97} & 38.19 & 44.34 & \multicolumn{1}{c}{44.13} & 25.88 & 27.56 & 31.17 \\
\multicolumn{1}{c}{Fusion with DSGN++} & 78.15 & 85.49 & \multicolumn{1}{c}{95.48} & 51.03 & 59.48 & \multicolumn{1}{c}{71.58} & 46.83 & 48.25 & 53.23 \\
\multicolumn{1}{c}{{\iffalse \cellcolor{lightgray!40} \fi \textit{Improvement}}} & {\iffalse \cellcolor{lightgray!40} \fi \textit{+29.15}} & {\iffalse \cellcolor{lightgray!40} \fi \textit{+18.66}} & \multicolumn{1}{c}{{\iffalse \cellcolor{lightgray!40} \fi \textit{+28.51}}} & {\iffalse \cellcolor{lightgray!40} \fi \textit{+12.84}} & {\iffalse \cellcolor{lightgray!40} \fi \textit{+15.14}} & \multicolumn{1}{c}{{\iffalse \cellcolor{lightgray!40} \fi \textit{+27.45}}} & {\iffalse \cellcolor{lightgray!40} \fi \textit{+20.95}} & {\iffalse \cellcolor{lightgray!40} \fi \textit{+20.69}} & {\iffalse \cellcolor{lightgray!40} \fi \textit{+22.06}} \\ \midrule
\multicolumn{1}{c}{16-LiDAR SECOND4x} & 65.31 & 78.28 & \multicolumn{1}{c}{80.18} & 52.97 & 58.62 & \multicolumn{1}{c}{59.40} & 43.86 & 47.37 & 49.35 \\
\multicolumn{1}{c}{Fusion with DSGN++} & 79.41 & 87.55 & \multicolumn{1}{c}{95.17} & 58.07 & 66.18 & \multicolumn{1}{c}{74.55} & 54.08 & 55.95 & 59.63 \\
\multicolumn{1}{c}{{\iffalse \cellcolor{lightgray!40} \fi \textit{Improvement}}} & {\iffalse \cellcolor{lightgray!40} \fi \textit{+14.10}} & {\iffalse \cellcolor{lightgray!40} \fi \textit{+9.27}} & \multicolumn{1}{c}{{\iffalse \cellcolor{lightgray!40} \fi \textit{+14.99}}} & {\iffalse \cellcolor{lightgray!40} \fi \textit{+5.10}} & {\iffalse \cellcolor{lightgray!40} \fi \textit{+7.56}} & \multicolumn{1}{c}{{\iffalse \cellcolor{lightgray!40} \fi \textit{+15.15}}} & {\iffalse \cellcolor{lightgray!40} \fi \textit{+10.22}} & {\iffalse \cellcolor{lightgray!40} \fi \textit{+8.58}} & {\iffalse \cellcolor{lightgray!40} \fi \textit{+10.28}} \\ \midrule
\multicolumn{1}{c}{64-LiDAR SECOND4x} & 81.23 & 89.52 & \multicolumn{1}{c}{94.37} & 59.40 & 62.71 & \multicolumn{1}{c}{68.50} & 61.90 & 62.45 & 68.87 \\
\multicolumn{1}{c}{Fusion with L-DSGN} & 81.13 & 88.36 & \multicolumn{1}{c}{94.50} & 60.05 & 62.65 & 68.42 & \multicolumn{1}{c}{56.94} & 57.05 & 65.84 \\
\multicolumn{1}{c}{Fusion with DSGN++} & \textbf{85.32} & \textbf{91.37} & \multicolumn{1}{c}{\textbf{95.79}} & \textbf{63.03} & \textbf{68.87} & \multicolumn{1}{c}{\textbf{76.72}} & \textbf{63.79} & \textbf{66.18} & \textbf{73.87} \\
\multicolumn{1}{c}{{\iffalse \cellcolor{lightgray!40} \fi \textit{Improvement}}} & {\iffalse \cellcolor{lightgray!40} \fi \textit{+4.09}} & {\iffalse \cellcolor{lightgray!40} \fi \textit{+1.85}} & \multicolumn{1}{c}{{\iffalse \cellcolor{lightgray!40} \fi \textit{+1.42}}} & {\iffalse \cellcolor{lightgray!40} \fi \textit{+3.63}} & {\iffalse \cellcolor{lightgray!40} \fi \textit{+6.16}} & \multicolumn{1}{c}{{\iffalse \cellcolor{lightgray!40} \fi \textit{+8.22}}} & {\iffalse \cellcolor{lightgray!40} \fi \textit{+1.89}} & {\iffalse \cellcolor{lightgray!40} \fi \textit{+3.73}} & {\iffalse \cellcolor{lightgray!40} \fi \textit{+5.00}} \\ \bottomrule
\end{tabular}
	\end{center}
\caption{\textbf{Performance comparison on the KITTI \textit{val} set in various modality settings.} Results in \textit{moderate} difficulty regime for all categories are provided as the main metric. $^\dag$ means training another model for \textsl{Pedestrian} and \textsl{Cyclist}. Best results are highlighted in \textbf{bold} for each sensor setup. {  \iffalse \color{red} \fi    LiDAR Sup. (L Sup.) represents whether to apply LiDAR depth supervision.} }
		\label{tab: KITTI val results}
\end{table*}

\subsubsection{Method Performance in Stereo Setup~~}

As shown in TABLE \ref{tab: KITTI val results}, we build our model based on two networks DSGN \cite{chen2020dsgn} (PSMNet \cite{psmnet} backbone) and L-DSGN \cite{guo2021liga} (ResNet-34 backbones). We adopt the same experimental settings for the respective experiments. By incorporating the proposed techniques, our method significantly surpasses the baselines: DSGN (\textsl{Car}: +\textbf{5.53} AP$_{3D}$, \textsl{Ped.}: +\textbf{8.78} AP$_{3D}$, \textsl{Cyc.}: +\textbf{11.67} AP$_{3D}$) and L-DSGN (\textsl{Car}: +\textbf{5.54} AP$_{3D}$, \textsl{Ped.}: +\textbf{8.32} AP$_{3D}$, \textsl{Cyc.}: +\textbf{16.19} AP$_{3D}$) regarding the moderate evaluation difficulty. The more noticeable improvements for \textsl{Pedestrian} and \textsl{Cyclist} reveal that the imbalanced class learning is greatly alleviated. {  \iffalse \color{red} \fi    Even without LiDAR signals supervision, our method still achieves 66+ AP$_{3D}$ for \textsl{Car} category and demonstrates significant improvements (\textsl{Car}: +\textbf{10.86} AP$_{3D}$, \textsl{Ped.}: +\textbf{6.75} AP$_{3D}$, \textsl{Cyc.}: +\textbf{17.15} AP$_{3D}$) compared with the baseline.}

\subsubsection{Method Performance in Multi-Modal Setup~~}

We further validate the effectiveness of our approaches in the multi-modal setting -- binocular cameras and LiDAR. The adopted LiDAR baseline network is SECOND4x \cite{second, guo2021liga}, which downsamples the sparse grid by four times on the bird's eye view and has the same grid size with stereo volume. Without special design, we simply fuse the learned Dual-view Stereo Volume by feature addition with LiDAR feature volume generated by SECOND4x. This multi-modal modeling uses the same training setup as in Sec.~\ref{sec: experimental setup}. 

We conduct a set of experiments that input LiDAR signals from sparse (4 beams) to dense (complete 64 beams) to validate the complementary effects of stereo cameras. The low-beams simulation of LiDAR signals follows \cite{pseudo++}.
As shown in TABLE \ref{tab: KITTI val results}, SECOND4x cannot handle well with the low-beams LiDAR and gets only $<50$ AP$_{3D}$ with inputs of $\leq 8$-beams LiDAR. With the simple fusion above with stereo features, all the LiDAR networks obtain noticeable accuracy gain consistently. The sparser the LiDAR signal is, the better the acquired improvements are. For example, an 8-beams LiDAR model gets an accuracy boost of $29.15$ AP$_{3D}$ to $78.15$ AP, which is even comparable to the reported results of SECOND with inputs of 64-beams LiDAR. {  \iffalse \color{red} \fi    Interestingly, DSGN++ with 4 beams even performs worse than DSGN++, which indicates direct fusion of extremely sparse LiDAR (4-beams) is potentially harmful to stereo 3D detectors. Compared with PV-RCNN, DSGN++ with 64 beams performs better in primary metric AP$_{3D}$ and AP$_{2D}$ while getting worse results in AP$_{BEV}$. The results reveal the LiDAR model benefits more from accurate front-view detection performance than the top-view one. As a result, the multi-modal 3D detector still performs better in 3D metrics.}

In particular for the complete 64-beams LiDAR setup, direct multi-modal modeling by fusing L-DSGN even deteriorates the detection performance. In contrast, multi-modal modeling with DSGN++ yet improves the LiDAR network significantly (\textsl{Car}: +\textbf{4.09} AP$_{3D}$, \textsl{Ped.}: +\textbf{3.63} AP$_{3D}$, \textsl{Cyc.}: +\textbf{1.89} AP$_{3D}$). This comparison reveals the fact that stereo representation can provide strong complementary cues over the vanilla LiDAR signals. The fusion of multi-sensor is promising and improves the robustness of 3D perception system. 

\subsection{Ablation Study}\label{sec: ablation study}

\begin{table*}[bpt]
	\begin{center}
\begin{tabular}{l|c|x{0.52in}x{0.52in}x{0.52in}cccccc}
\toprule
\multicolumn{1}{c}{\multirow{2}{*}{id.}} & \multicolumn{1}{c}{\multirow{2}{*}{Pipelines}} & \multicolumn{1}{c}{\multirow{2}{*}{FSD Head}} & \multicolumn{1}{c}{\multirow{2}{*}{D-PS}} & \multicolumn{1}{c}{\multirow{2}{*}{SLCP}} & \multicolumn{2}{c}{\textsl{Car} } & \multicolumn{2}{c}{\textsl{Pedestrian}   } & \multicolumn{2}{c}{\textsl{Cyclist} } \\ \cmidrule(lr){6-7} \cmidrule(lr){8-9} \cmidrule(lr){10-11}
\multicolumn{1}{c}{} & \multicolumn{1}{c}{} &  &  &  & AP$_{3D}$ & AP$_{BEV}$ & AP$_{3D}$ & AP$_{BEV}$ & AP$_{3D}$ & AP$_{BEV}$ \\ \midrule
a. & \multirow{3}{*}{Front-View} & \checkmark & & \multicolumn{1}{c|}{} & 63.58 & 73.53 & 33.12 & 40.50 & 28.09 & 29.58 \\ 
b. & & \checkmark & \checkmark & \multicolumn{1}{c|}{} & 66.42 & 77.13 & 34.91 & 41.76 & 30.17 & 34.29 \\
c. &  & \checkmark & \checkmark & \multicolumn{1}{c|}{\checkmark} & 67.63 & 76.73 & \textbf{42.98} & \textbf{50.64} & 41.63 & 43.96 \\ \midrule
d. & \multirow{4}{*}{Top-View} & & & \multicolumn{1}{c|}{} & 56.69 & 66.05 & 28.23 & 32.08 & 18.59 & 17.47 \\ 
e. &  & \checkmark &  & \multicolumn{1}{c|}{}
& 61.33 & 71.45 & 29.14 & 35.02 & 20.51 & 19.71 \\ 
f. &  & \checkmark & \checkmark & \multicolumn{1}{c|}{} & 64.59 & 74.14 & 29.38 & 39.12 & 20.45 & 21.28 \\
g. & & \checkmark & \checkmark & \multicolumn{1}{c|}{\checkmark} & 67.85 & 77.16 & 36.20 & 43.09 & 29.17 & 30.56 \\ \midrule
h. & \multirow{4}{*}{Dual-View} & & & \multicolumn{1}{c|}{} & 61.46 & 67.81 & 33.41 & 40.59 & 23.42 & 24.74 \\
i. &  & \checkmark &  & \multicolumn{1}{c|}{} & 64.43 & 74.19 & 33.51 & 42.48 & 33.82 & 35.64 \\
j. &  & \checkmark & \checkmark & \multicolumn{1}{c|}{} & 66.21 & 76.96 & 37.19 & 44.50 & 30.01 & 32.69 \\
k. &  & \checkmark & \checkmark & \multicolumn{1}{c|}{\checkmark} & \textbf{69.12} & \textbf{78.93} & 42.44 & 50.06 & \textbf{42.48} & \textbf{45.77} \\ \bottomrule
\end{tabular}
	\end{center}
\caption{\textbf{Main ablation studies on the KITTI \textit{val} set.}
    As illustrated in Fig.~\ref{fig: different stereo volumes}, we separate the pipelines according to the information flow types -- front-view (FV), top-view (TV), and dual-view (DV).
    FSD Head denotes the application of the front-surface depth head (Sec.~\ref{front-surface head}) for geometric learning. SLCP represents \textit{Stereo-LiDAR copy-paste}.  Originally FV applies depth head and TV is supervised by voxel occupancy loss. ``--'' denotes the component that is not applicable in the respective pipeline.}
		\label{tab: Main Ablation Study}
\end{table*}

In this section, we investigate the effectiveness of the major adaptations. For fair comparisons, we conduct ablation studies on the KITTI \textit{val} set mainly in TABLE \ref{tab: Main Ablation Study}. Note that we primarily adopt the accuracy of \textsl{Car} category for the ablation study by default, as the dataset contains fewer annotations for \textsl{Pedestrian} and \textsl{Cyclist} that causes greater result variance.


\subsubsection{Ablation Study for Depth-wise Plane Sweeping~~}\label{sec: ablation study}

As shown in TABLE \ref{tab: Main Ablation Study} (a. \textit{vs.} b.; d. \textit{vs.} e.; h. \textit{vs.} i.), with similar computations, models with depth-wise plane sweeping obtains extra 1.5$\sim$3.2 AP improvements. In general, performance gains are consistent for all three categories, which indicates that traditional plane sweeping is unsuited for representing complicated predictions, and D-PS much eases it. 

TABLE \ref{ablation for dps} ablates the expanded channels $C_I$ and smoothness factor $\alpha$ for plane sweeping. We observe that the wider 2D features contribute to the learning of 2D semantic features. We compare D-PS with another possible choice (called Group-PS) to input the features with expanded channels: equally splitting the channels/depths into several groups; extracting the depth-wise features in respective groups. However, Group-PS cannot guarantee feature's channel-wise similarity between groups for stereo matching and yields $65.48$ AP$_{3D}$. D-PS is designed to preserve the local feature \textit{continuity} and the degree of sharing can be adjusted w.r.t disparity. The smoothing factors $\alpha=0.1$ ($66.42$ AP$_{3D}$) for FV and $\alpha=0.5$ ($64.59$ AP$_{3D}$) for TV yield best performance. Dual-view stereo volume adopts the respective best-performing parameters for both volumes.

In addition, to illustrate the generality of D-PS, we also provide the monocular experiments. We simply adapt the top-view pipeline for monocular 3D detectors (remove right image input and right feature) and keep other training setups the same. As shown at the bottom of TABLE \ref{ablation for dps}, the method with D-PS surpasses the baseline by 2.05 AP$_{3D}$ and 2.81 AP$_{BEV}$. 

\begin{table}[bpt]
    \centering
    \renewcommand{\arraystretch}{1.08}
    \begin{tabular}{c|ccc|cc}
        \toprule
        \multicolumn{1}{c}{\multirow{2}{*}{Pipelines}} & \multicolumn{1}{c}{\multirow{2}{*}{Sampling}} & \multicolumn{1}{c}{\multirow{2}{*}{\#Chn}} & \multicolumn{1}{c}{\multirow{2}{*}{$\alpha$}} & \multicolumn{2}{c}{Car} \\ \cmidrule{5-6}
        \multicolumn{1}{c}{}& \multicolumn{1}{c}{} & \multicolumn{1}{c}{} & \multicolumn{1}{c}{} & \multicolumn{1}{c}{AP$_{3D}$} & AP$_{BEV}$ \\ \midrule
        \multicolumn{6}{l}{\ \ \textit{Stereo Camera Sensor}} \\ \midrule
        \multirow{6}{*}{FV} & PS & 32 & -- & \multicolumn{1}{c}{63.58} & 73.53 \\ 
        & Group-PS & 96 & 32-sep & \multicolumn{1}{c}{64.98} & 76.02 \\ 
        & & 48 & 0.1 & \multicolumn{1}{c}{65.48} & 76.28 \\ 
        & & 96 & 1. & \multicolumn{1}{c}{65.95} & 76.71 \\ 
        & & 96 & 0.1 & \multicolumn{1}{c}{\textbf{66.42}} & \textbf{77.13} \\ 
        & \multirow{-4}{*}{\iffalse \cellcolor{lightgray!40} \fi D-PS} & 96 & 0.01 & \multicolumn{1}{c}{66.28} & 76.82 \\ \midrule
        \multirow{6}{*}{TV} & PS & 32 & -- & \multicolumn{1}{c}{61.33} & 71.45 \\ 
        & Group-PS & 96 & 32-sep & \multicolumn{1}{c}{62.50} & 72.60 \\ 
        & & 48 & 0.1 & \multicolumn{1}{c}{62.59} & 73.22 \\ 
        & & 96 & 1 & \multicolumn{1}{c}{63.85} & 73.59 \\ 
        & & 96 & 0.5 & \multicolumn{1}{c}{\textbf{64.59}} & \textbf{74.14} \\ 
        & \multirow{-4}{*}{\iffalse \cellcolor{lightgray!40} \fi D-PS} & 96 & 0.1 & \multicolumn{1}{c}{64.15} & 73.40 \\ \midrule
        \multicolumn{6}{l}{\ \ \textit{Monocular Camera Sensor}} \\ \midrule
        \multirow{2}{*}{TV} & PS & 32 & -- & 15.36 & 21.23 \\ 
         & D-PS & 96 & 0.1 & \textbf{17.41} & \textbf{24.04} \\ \midrule
    \end{tabular}
    \caption{\textbf{Effects of expanded channels and smoothness factor $\alpha$ for Depth-wise Plane Sweeping.} \#Chn denotes the channel number $C_I$ of 2D feature for building volumes. ``Group-PS'' (32-sep) represents sweeping planes by propagates features from the evenly-spaced channel groups (each with 32 channels) to the depth planes of the respective group. }
    \label{ablation for dps}
\end{table}

\subsubsection{Ablation Study for Stereo Volumetric Representation}

We conduct the comparison of several stereo pipelines (See information flows in Fig. \ref{fig: different stereo volumes})). 

\noindent \textbf{Effects of Front-Surface Depth Head.~~}
TABLE \ref{tab: Main Ablation Study}(d. \textit{vs.} e., h. \textit{vs.} i.) compares the effects of different depth supervision signals (voxel occupancy head vs. front-surface depth head), where FSD head yields the respective performance gains of \textsl{Car}: +4.64 AP$_{3D}$ and +2.97 AP$_{3D}$. The improvement demonstrates that FSD head can take in more precise depth signals than discretized occupancy classification. We conjecture the consistent shape between cost volume and input views assists sparse depth supervision with front-view context features between sparse beams.

\noindent \textbf{Effects of Volume Integration.~~} 
For fair comparisons of both single-view representations, we ensure the front-view pipeline and top-view one share similar volume sizes: PSV is of shape $72\times 80\times 32 = 1797120$ and 3DGV has the shape of $20\times 304\times 288 = 1751040$, and there is only $2.6\%$ calculation counts difference. As shown in TABLE \ref{tab: Main Ablation Study} (c. \textit{vs.} g.; b. \textit{vs.} f.), despite the input of balanced foreground categories, TV branch cannot achieve the same performance for \textsl{Ped.} and \textsl{Cyc.}, which shows top-view representation may not be well-suited for smaller objects in bird's eye view. The volume integration (k. in TABLE \ref{tab: Main Ablation Study}) boosts the performance to (69.12 AP$_{3D}$, 78.93 AP$_{BEV}$) for \textsl{Car}, and (42.48 AP$_{3D}$, 45.77 AP$_{BEV}$) for \textsl{Cyc.}. In comparison of c., g. and k., top-view representation provides more complementary cues for \textsl{Car} and \textsl{Cyc.}.

\subsubsection{Ablation Study for Stereo-LiDAR Copy-Paste}


\begin{table}[bpt]
    \centering
    \begin{tabular}{l|cl|ccc}
        \toprule
        \multicolumn{1}{c}{id.} & \multicolumn{1}{c}{Samples} & \multicolumn{1}{c}{Prob} & \textsl{Car} & \textsl{Ped.} & \textsl{Cyc.} \\ \midrule
        a. & -- & 0. & 66.21 & 37.19 & 30.01 \\ 
        b. & \{3, 3, 3\} & 0.6 & 67.79 & 39.66 & 39.56 \\ 
        c. & \{5, 0, 0\} & 0.6 & 68.74 & 32.50 & 28.81 \\ 
        d. & \{0, 5, 5\} & 0.6 & 66.18 & 39.42 & 43.22 \\ 
        e. & & 0.4 & 68.79 & 39.20 & \textbf{43.51} \\ 
        f. & & 0.6 & \textbf{69.12} & \textbf{42.44} & 42.48 \\  
        g. & & 0.8 & 68.41 & 41.74 & 39.47 \\ 
        h. & \multirow{-4}{*}{ \iffalse \cellcolor{lightgray!40} \fi \{5, 5, 5\} } & 1. & 68.55 & 40.18 & 38.58 \\ 
        i. & \{7, 7, 7\} & 0.6 & 68.92 & 41.03 & 42.93 \\ 
        j. & \multicolumn{2}{c|}{\iffalse \cellcolor{lightgray!40} \fi w/o Occ Removal} & 66.52 & 38.79 & 36.27 \\ \bottomrule
   \end{tabular}
    \caption{\textbf{Hyper-parameter choices for \textit{Stereo-LiDAR copy-paste}.} Models are evaluated using AP$_{3D}$ (Moderate) on the KITTI \textit{val} set. ``Samples'' denotes the augmented object counts for \textsl{Car}, \textsl{Pedestrian}, and \textsl{Cyclist}, respectively. ``Prob'' indicates the apply SLCP for each training scene. ``w/o Occ Removal'' denotes cancelling the removal of occluded point clouds. ``Paste by Distance'' represents pasting objects from near to far. }
    \label{ablation for stereo lidar}
\end{table}

TABLE \ref{tab: Main Ablation Study} (j. \textit{vs.} k.) shows that SLCP augmentation improves data efficiency (\textsl{Car}: +3.96 AP$_{3D}$) and greatly mitigates imbalanced class learning (\textsl{Ped.}:  +5.25 AP$_{3D}$, \textsl{Cyc.} +12.47 AP$_{3D}$.

\begin{figure*}[htbp]
    \begin{center}
    \includegraphics[width=185mm]{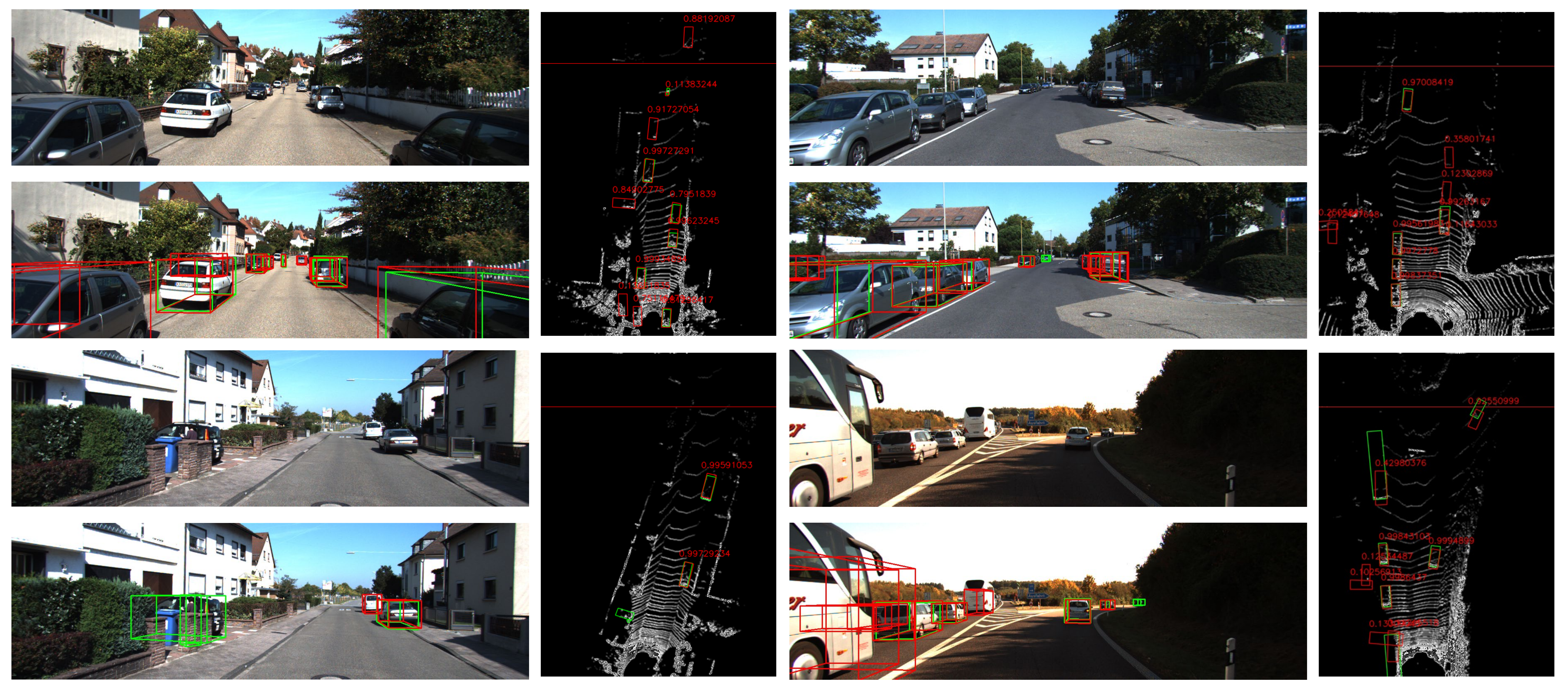} 
    \end{center}
    \caption{\textbf{Qualitative results on the KITTI \textit{val} set.} Green boxes represents ground-truth and red boxes denotes our predictions. The left-view images are shown in the left column and the BEV point clouds images are shown on the right side. Some failure cases are shown at the bottom of the table. Please zoom in to observe the prediction details. Redline shown in the bird's eye view is 50 meters away from the sensor.}
    \label{fig:good result}
\end{figure*}

\begin{figure}[bpt]
    \begin{center}
    \includegraphics[width=90mm]{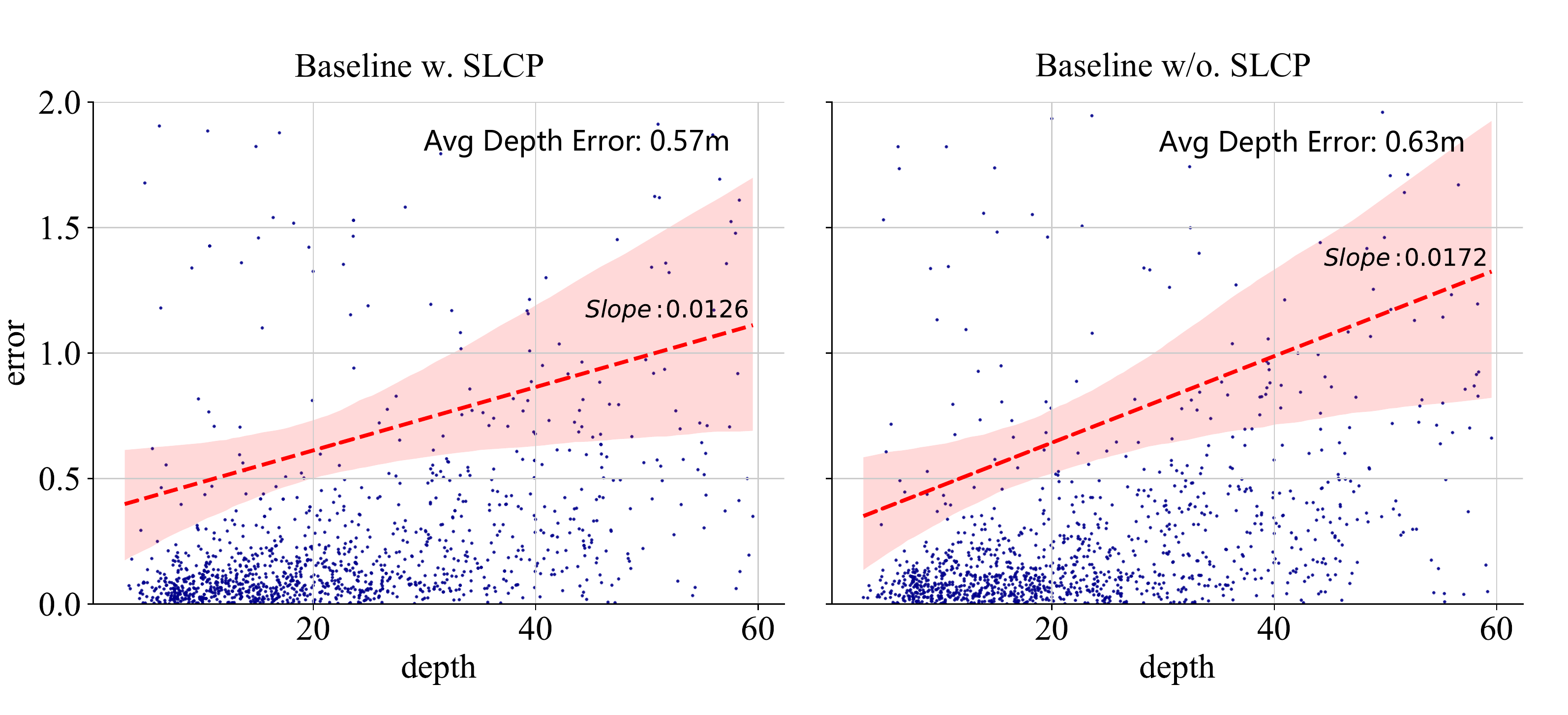}
    \end{center}
    \caption{\textbf{Comparison of foreground object localization error. Red dotted line computes the regressed line.}  The localization error computes the average depth estimation error within the 3D object boxes at the respective depth ranges. }
    \label{fig:depth plot}
\end{figure}

In terms of depth estimation quality, as visualized in Fig. \ref{fig:depth plot}, the ratio of foreground localization error w.r.t object distances gets a smoother slope. Overall, our model reduces the foreground depth estimation error from $0.63$ to $0.57$ ($m$). 

TABLE \ref{ablation for stereo lidar} ablates several hyper-parameters used in \textit{Stereo-LiDAR copy-paste}. TABLE \ref{ablation for stereo lidar} (a., b., f., i.) ablates the number of pasted samples into the training scenes and 5 samples yield the best performance. {  \iffalse \color{red} \fi    Compared with raw data pair inputs (a.), the sufficient pasted objects greatly alleviate the imbalanced problems across categories (f.).}
{  \iffalse \color{red} \fi    TABLE \ref{ablation for stereo lidar} (a., c., d.) ablates the effect of more positive numbers, where the improvement of $+2.53$ AP$_{3D}$ for \textsl{Car} (TABLE \ref{ablation for stereo lidar} (a. \textit{vs.} c.)) indicates that the current positive instances limit the modeling efficiency even for the most frequent class.} (e. - h.) ablates the probability of applying copy-paste. As the image-level copy-paste produces object patch artifacts and occlusion, the larger probability can lead to worse results. TABLE \ref{ablation for stereo lidar}(f. \textit{vs.} j.) ablates whether to remove the background point clouds and the results reveal the removal of background points facilitates the model to better learn from copy-paste augmentation (\textsl{Car}: $+2.60$ AP$_{3D}$ ). 

\subsection{Qualitative Results} \label{sec: visualization results}

We present some representative results in Fig.~\ref{fig:good result}, especially for the occlusion situations. From the visualization of BEV results, our approach robustly predicts most objects and estimates their accurate 3D bounding boxes even for the scene 50 meters away (red line in the bird's eye view). The visualization shows the great potential for the low-cost outdoor perception system based on stereo cameras. Noticeably, some extremely occluded cases in the top row could be still detected. The bottom row in Fig.~\ref{fig:good result} also visualizes some failure cases including missing occluded objects, missing distant objects, and wrong orientation/dimension predictions.

\begin{table}[bpt]
\begin{center}
\begin{tabular}{cccc}
\toprule
 & \textsl{Car} AP$_{3D}$ & \textsl{Car} AP$_{BEV}$ & Inference Time \\ \midrule
DSGN++ & \textbf{69.12} & \textbf{78.93} & 0.281s \\
{  \iffalse \color{red} \fi    FV-DSGN++} & {  \iffalse \color{red} \fi    67.63} & {  \iffalse \color{red} \fi    76.73} & {  \iffalse \color{red} \fi    0.198s} \\
{  \iffalse \color{red} \fi    TV-DSGN++} & {  \iffalse \color{red} \fi    67.85} & {  \iffalse \color{red} \fi    77.16} & {  \iffalse \color{red} \fi    0.202s} \\
R18-DSGN++ & 68.12 & 77.01 & \textbf{0.178}s \\
\bottomrule
\end{tabular}
\end{center}
\caption{\textbf{Inference time comparison and the efficient implementation.} The experiments are conducted on a NVIDIA \textit{RTX 2080Ti} GPU with batch size of 1. } \label{tab: speed}
\end{table}

\subsection{Efficiency Study} \label{sec: speed}

As the large performance gap between LiDAR-based approaches and camera-based approaches \cite{CaDDN, plume, guo2021liga}, most works focus on the improvement of detection accuracy. Efficiency comparison of different algorithms is less investigated due to various experimental setups.

For a fair comparison, we conduct the efficiency comparisons on an NVIDIA \textit{RTX 2080TI} GPU as shown in TABLE \ref{tab: speed}. Generally, the complete DSGN++ with ResNet-34 runs takes $0.273$s on average, where binocular feature extraction takes $2\times 0.058 = 0.116$s, PSV takes $0.036$s, 3DGV takes $0.045$s, DSV and 3D network take $0.044$s and last BEV detector costs $0.012$s. {  \iffalse \color{red} \fi    The inference time of single-view pipelines TV-DSGN++ and FV-DSGN++ are 0.198s and 0.202s, respectively. Despite their lower performance compared with DSGN++, the simple single-view pipelines can also serve as simple baselines for stereo 3D object detection for faster speed.}

{  \iffalse \color{red} \fi    For accelerating our pipeline, we provide another efficient implementation (R18-DSGN++) to demonstrate the speed-accuracy trade-off of our work.} We replace ResNet-34 with ResNet-18 and adopt the same 2D upsampling head for both stereo volumes. The efficient model still achieves 68.12 AP$_{3D}$ despite the backbone network removing about half of the parameters. This fact also indicates that the original 2D backbone network is not fully exploited for the construction of the following stereo volumes as described in Sec. \ref{sec: depth-wise plane sweeping}. 

We note that the code is not yet fully optimized and affects the speed of the 3D detector. For example, despite the same computation, the CUDA implementation of D-PS costs extra 4ms than PyTorch built-in implementation of \textit{F.grid\_sample} for plane sweeping. We leave the further code optimization to future work.

\section{Conclusion}
We have renewed several key components that build an end-to-end stereo detection pipeline and provided a new stereo modeling -- DSGN++ -- for 3D object detection. Without bells and whistles, we conducted a set of comprehensive experiments to illustrate the effectiveness of the proposed modules. Specifically, the proposed depth-wise plane sweeping allows inputs of wider 2D features and improves modeling efficiency in 2D-to-3D transformation. Dual-view stereo volumes provide better 3D representations that grasp differently spaced features. And \textit{Stereo-LiDAR copy-paste} strategy largely improves data efficiency and enhances modeling generalization ability for \textit{all} categories. We expect the framework provides a strong baseline for the future application of camera-based 3D perception systems.


%





\ifCLASSOPTIONcaptionsoff
  \newpage
\fi



\bibliographystyle{IEEEtran}
\bibliography{egbib}

\begin{thebibliography}{10}
\providecommand{\url}[1]{#1}
\csname url@samestyle\endcsname
\providecommand{\newblock}{\relax}
\providecommand{\bibinfo}[2]{#2}
\providecommand{\BIBentrySTDinterwordspacing}{\spaceskip=0pt\relax}
\providecommand{\BIBentryALTinterwordstretchfactor}{4}
\providecommand{\BIBentryALTinterwordspacing}{\spaceskip=\fontdimen2\font plus
\BIBentryALTinterwordstretchfactor\fontdimen3\font minus
  \fontdimen4\font\relax}
\providecommand{\BIBforeignlanguage}[2]{{%
\expandafter\ifx\csname l@#1\endcsname\relax
\typeout{** WARNING: IEEEtran.bst: No hyphenation pattern has been}%
\typeout{** loaded for the language `#1'. Using the pattern for}%
\typeout{** the default language instead.}%
\else
\language=\csname l@#1\endcsname
\fi
#2}}
\providecommand{\BIBdecl}{\relax}
\BIBdecl

\bibitem{rcnn}
R.~Girshick, J.~Donahue, T.~Darrell, and J.~Malik, ``Rich feature hierarchies
  for accurate object detection and semantic segmentation,'' in \emph{CVPR},
  2014, pp. 580--587.

\bibitem{fastrcnn}
R.~Girshick, ``Fast r-cnn,'' in \emph{ICCV}, 2015.

\bibitem{fasterrcnn}
S.~Ren, K.~He, R.~Girshick, and J.~Sun, ``Faster r-cnn: Towards real-time
  object detection with region proposal networks,'' in \emph{NeurIPS}, 2015.

\bibitem{ssd}
W.~Liu, D.~Anguelov, D.~Erhan, C.~Szegedy, S.~Reed, C.-Y. Fu, and A.~C. Berg,
  ``Ssd: Single shot multibox detector,'' in \emph{ECCV}, 2016.

\bibitem{dorn}
H.~Fu, M.~Gong, C.~Wang, K.~Batmanghelich, and D.~Tao, ``Deep ordinal
  regression network for monocular depth estimation,'' in \emph{CVPR}, 2018,
  pp. 2002--2011.

\bibitem{psmnet}
J.-R. Chang and Y.-S. Chen, ``Pyramid stereo matching network,'' in
  \emph{CVPR}, 2018, pp. 5410--5418.

\bibitem{gcnet}
A.~Kendall, H.~Martirosyan, S.~Dasgupta, P.~Henry, R.~Kennedy, A.~Bachrach, and
  A.~Bry, ``End-to-end learning of geometry and context for deep stereo
  regression,'' in \emph{ICCV}, 2017, pp. 66--75.

\bibitem{stereorcnn}
P.~Li, X.~Chen, and S.~Shen, ``Stereo r-cnn based 3d object detection for
  autonomous driving,'' in \emph{CVPR}, 2019, pp. 7644--7652.

\bibitem{qin2019monogrnet}
Z.~Qin, J.~Wang, and Y.~Lu, ``Monogrnet: A geometric reasoning network for
  monocular 3d object localization,'' in \emph{AAAI}, vol.~33, 2019, pp.
  8851--8858.

\bibitem{MLF}
B.~Xu and Z.~Chen, ``Multi-level fusion based 3d object detection from
  monocular images,'' in \emph{CVPR}, 2018, pp. 2345--2353.

\bibitem{zhang2021objects}
Y.~Zhang, J.~Lu, and J.~Zhou, ``Objects are different: Flexible monocular 3d
  object detection,'' in \emph{CVPR}, 2021, pp. 3289--3298.

\bibitem{ding2020learning}
M.~Ding, Y.~Huo, H.~Yi, Z.~Wang, J.~Shi, Z.~Lu, and P.~Luo, ``Learning
  depth-guided convolutions for monocular 3d object detection,'' in \emph{CVPR
  Workshops}, 2020, pp. 1000--1001.

\bibitem{triangulation}
Z.~Qin, J.~Wang, and Y.~Lu, ``Triangulation learning network: from monocular to
  stereo 3d object detection,'' \emph{CVPR}, 2019.

\bibitem{disentangling}
A.~Simonelli, S.~Rota~Bulo, L.~Porzi, M.~Lopez-Antequera, and P.~Kontschieder,
  ``Disentangling monocular 3d object detection,'' in \emph{ICCV}, 2019.

\bibitem{brazil2019m3d}
G.~Brazil and X.~Liu, ``M3d-rpn: Monocular 3d region proposal network for
  object detection,'' in \emph{arXiv preprint arXiv:1907.06038}, 2019.

\bibitem{fcos3d}
T.~Wang, X.~Zhu, J.~Pang, and D.~Lin, ``Fcos3d: Fully convolutional one-stage
  monocular 3d object detection,'' \emph{arXiv preprint arXiv:2104.10956},
  2021.

\bibitem{oftnet}
T.~Roddick, A.~Kendall, and R.~Cipolla, ``Orthographic feature transform for
  monocular 3d object detection,'' in \emph{British Machine Vision Conference},
  2019.

\bibitem{pseudolidar}
Y.~Wang, W.-L. Chao, D.~Garg, B.~Hariharan, M.~Campbell, and K.~Q. Weinberger,
  ``Pseudo-lidar from visual depth estimation: Bridging the gap in 3d object
  detection for autonomous driving,'' in \emph{CVPR}, 2019, pp. 8445--8453.

\bibitem{chen2020dsgn}
Y.~Chen, S.~Liu, X.~Shen, and J.~Jia, ``Dsgn: Deep stereo geometry network for
  3d object detection,'' in \emph{CVPR}, 2020, pp. 12\,536--12\,545.

\bibitem{accuratemono3d}
X.~Ma, Z.~Wang, H.~Li, W.~Ouyang, and P.~Zhang, ``Accurate monocular 3d object
  detection via color-embedded 3d reconstruction for autonomous driving,'' in
  \emph{arXiv preprint arXiv:1903.11444}, 2019.

\bibitem{pseudo++}
Y.~You, Y.~Wang, W.-L. Chao, D.~Garg, G.~Pleiss, B.~Hariharan, M.~Campbell, and
  K.~Q. Weinberger, ``Pseudo-lidar++: Accurate depth for 3d object detection in
  autonomous driving,'' in \emph{ICLR}, 2020.

\bibitem{pointrcnn}
S.~Shi, X.~Wang, and H.~Li, ``Pointrcnn: 3d object proposal generation and
  detection from point cloud,'' in \emph{CVPR}, 2019, pp. 770--779.

\bibitem{second}
Y.~Yan, Y.~Mao, and B.~Li, ``Second: Sparsely embedded convolutional
  detection,'' in \emph{Sensors}, 2018.

\bibitem{VoxelNet}
Y.~Zhou and O.~Tuzel, ``Voxelnet: End-to-end learning for point cloud based 3d
  object detection,'' in \emph{CVPR}, 2018.

\bibitem{plume}
Y.~Wang, B.~Yang, R.~Hu, M.~Liang, and R.~Urtasun, ``Plume: Efficient 3d object
  detection from stereo images,'' in \emph{2020 IEEE/RSJ International
  Conference on Intelligent Robots and Systems (IROS)}, 2021.

\bibitem{endtoendpseudolidar}
R.~Qian, D.~Garg, Y.~Wang, Y.~You, S.~Belongie, B.~Hariharan, M.~Campbell,
  K.~Q. Weinberger, and W.-L. Chao, ``End-to-end pseudo-lidar for image-based
  3d object detection,'' in \emph{CVPR}, 2020, pp. 5881--5890.

\bibitem{CaDDN}
C.~Reading, A.~Harakeh, J.~Chae, and S.~L. Waslander, ``Categorical depth
  distributionnetwork for monocular 3d object detection,'' \emph{CVPR}, 2021.

\bibitem{guo2021liga}
X.~Guo, S.~Shi, X.~Wang, and H.~Li, ``Liga-stereo: Learning lidar geometry
  aware representations for stereo-based 3d detector,'' in \emph{ICCV}, 2021,
  pp. 3153--3163.

\bibitem{planesweep}
R.~T. Collins, ``A space-sweep approach to true multi-image matching,'' in
  \emph{CVPR}.\hskip 1em plus 0.5em minus 0.4em\relax IEEE, 1996, pp. 358--363.

\bibitem{deepstereo}
J.~Flynn, I.~Neulander, J.~Philbin, and N.~Snavely, ``Deepstereo: Learning to
  predict new views from the world's imagery,'' in \emph{CVPR}, 2016, pp.
  5515--5524.

\bibitem{mvsnet}
Y.~Yao, Z.~Luo, S.~Li, T.~Fang, and L.~Quan, ``Mvsnet: Depth inference for
  unstructured multi-view stereo,'' in \emph{ECCV}, 2018, pp. 767--783.

\bibitem{copypaste}
G.~Ghiasi, Y.~Cui, A.~Srinivas, R.~Qian, T.-Y. Lin, E.~D. Cubuk, Q.~V. Le, and
  B.~Zoph, ``Simple copy-paste is a strong data augmentation method for
  instance segmentation,'' in \emph{CVPR}, 2021, pp. 2918--2928.

\bibitem{kitti}
A.~Geiger, P.~Lenz, and R.~Urtasun, ``Are we ready for autonomous driving? the
  kitti vision benchmark suite,'' in \emph{CVPR}, 2012.

\bibitem{AVOD}
J.~Ku, M.~Mozifian, J.~Lee, A.~Harakeh, and S.~Waslander, ``Joint 3d proposal
  generation and object detection from view aggregation,'' in \emph{IROS},
  2018.

\bibitem{ganet}
F.~Zhang, V.~Prisacariu, R.~Yang, and P.~H. Torr, ``Ga-net: Guided aggregation
  net for end-to-end stereo matching,'' in \emph{CVPR}, 2019, pp. 185--194.

\bibitem{groupwisestereo}
X.~Guo, K.~Yang, W.~Yang, X.~Wang, and H.~Li, ``Group-wise correlation stereo
  network,'' in \emph{CVPR}, 2019.

\bibitem{pwcnet}
D.~Sun, X.~Yang, M.-Y. Liu, and J.~Kautz, ``Pwc-net: Cnns for optical flow
  using pyramid, warping, and cost volume,'' in \emph{CVPR}, 2018, pp.
  8934--8943.

\bibitem{anytimestereo}
Y.~Wang, Z.~Lai, G.~Huang, B.~H. Wang, L.~van~der Maaten, M.~Campbell, and
  K.~Q. Weinberger, ``Anytime stereo image depth estimation on mobile
  devices,'' in \emph{ICRA}.\hskip 1em plus 0.5em minus 0.4em\relax IEEE, 2019,
  pp. 5893--5900.

\bibitem{dispnet}
N.~Mayer, E.~Ilg, P.~Hausser, P.~Fischer, D.~Cremers, A.~Dosovitskiy, and
  T.~Brox, ``A large dataset to train convolutional networks for disparity,
  optical flow, and scene flow estimation,'' in \emph{CVPR}, 2016, pp.
  4040--4048.

\bibitem{hierarchical}
Z.~Yin, T.~Darrell, and F.~Yu, ``Hierarchical discrete distribution
  decomposition for match density estimation,'' in \emph{CVPR}, 2019, pp.
  6044--6053.

\bibitem{segstereo}
G.~Yang, H.~Zhao, J.~Shi, Z.~Deng, and J.~Jia, ``Segstereo: Exploiting semantic
  information for disparity estimation,'' in \emph{ECCV}, 2018, pp. 636--651.

\bibitem{learndispconsistency}
Z.~Liang, Y.~Feng, Y.~Guo, H.~Liu, W.~Chen, L.~Qiao, L.~Zhou, and J.~Zhang,
  ``Learning for disparity estimation through feature constancy,'' in
  \emph{CVPR}, 2018, pp. 2811--2820.

\bibitem{edgestereo}
X.~Song, X.~Zhao, H.~Hu, and L.~Fang, ``Edgestereo: A context integrated
  residual pyramid network for stereo matching,'' in \emph{ACCV}, 2018.

\bibitem{MVSMachine}
A.~Kar, C.~H{\"a}ne, and J.~Malik, ``Learning a multi-view stereo machine,'' in
  \emph{Advances in neural information processing systems}, 2017, pp. 365--376.

\bibitem{r-mvset}
Y.~Yao, Z.~Luo, S.~Li, T.~Shen, T.~Fang, and L.~Quan, ``Recurrent mvsnet for
  high-resolution multi-view stereo depth inference,'' in \emph{CVPR}, 2019.

\bibitem{point-mvsnet}
R.~Chen, S.~Han, J.~Xu, and H.~Su, ``Point-based multi-view stereo network,''
  in \emph{ICCV}, 2019.

\bibitem{surfacenet}
M.~Ji, J.~Gall, H.~Zheng, Y.~Liu, and L.~Fang, ``Surfacenet: An end-to-end 3d
  neural network for multiview stereopsis,'' in \emph{ICCV}, 2017, pp.
  2307--2315.

\bibitem{deepmvs}
P.-H. Huang, K.~Matzen, J.~Kopf, N.~Ahuja, and J.-B. Huang, ``Deepmvs: Learning
  multi-view stereopsis,'' in \emph{CVPR}, 2018, pp. 2821--2830.

\bibitem{casmvsnet}
X.~Gu, Z.~Fan, S.~Zhu, Z.~Dai, F.~Tan, and P.~Tan, ``Cascade cost volume for
  high-resolution multi-view stereo and stereo matching,'' in \emph{CVPR},
  2020.

\bibitem{MV3D}
X.~Chen, H.~Ma, J.~Wan, B.~Li, and T.~Xia, ``Multi-view 3d object detection
  network for autonomous driving,'' in \emph{CVPR}, 2017.

\bibitem{pointpillar}
A.~H. Lang, S.~Vora, H.~Caesar, L.~Zhou, J.~Yang, and O.~Beijbom,
  ``Pointpillars: Fast encoders for object detection from point clouds,'' in
  \emph{arXiv:1812.05784}, 2018.

\bibitem{fastpointrcnn}
Y.~Chen, S.~Liu, X.~Shen, and J.~Jia, ``Fast point r-cnn,'' in \emph{ICCV},
  2019, pp. 9775--9784.

\bibitem{pointnet}
C.~R. Qi, H.~Su, K.~Mo, and L.~J. Guibas, ``Pointnet: Deep learning on point
  sets for 3d classification and segmentation,'' in \emph{CVPR}, 2017.

\bibitem{pointnet++}
C.~R. Qi, L.~Yi, H.~Su, and L.~J. Guibas, ``Pointnet++: Deep hierarchical
  feature learning on point sets in a metric space,'' in \emph{NeurIPS}, 2017.

\bibitem{std}
Z.~Yang, Y.~Sun, S.~Liu, X.~Shen, and J.~Jia, ``Std: Sparse-to-dense 3d object
  detector for point cloud,'' in \emph{ICCV}, 2019.

\bibitem{pointfusion}
D.~Xu, D.~Anguelov, and A.~Jain, ``Pointfusion: Deep sensor fusion for 3d
  bounding box estimation,'' in \emph{CVPR}, 2018.

\bibitem{monopair}
Y.~Chen, L.~Tai, K.~Sun, and M.~Li, ``Monopair: Monocular 3d object detection
  using pairwise spatial relationships,'' in \emph{CVPR}, 2020, pp.
  12\,093--12\,102.

\bibitem{shi2021geometry}
X.~Shi, Q.~Ye, X.~Chen, C.~Chen, Z.~Chen, and T.-K. Kim, ``Geometry-based
  distance decomposition for monocular 3d object detection,'' \emph{arXiv
  preprint arXiv:2104.03775}, 2021.

\bibitem{shi2020distance}
X.~Shi, Z.~Chen, and T.-K. Kim, ``Distance-normalized unified representation
  for monocular 3d object detection,'' in \emph{European Conference on Computer
  Vision}.\hskip 1em plus 0.5em minus 0.4em\relax Springer, 2020, pp. 91--107.

\bibitem{centernet}
X.~Zhou, D.~Wang, and P.~Kr{\"a}henb{\"u}hl, ``Objects as points,'' in
  \emph{arXiv preprint arXiv:1904.07850}, 2019.

\bibitem{ma2021delving}
X.~Ma, Y.~Zhang, D.~Xu, D.~Zhou, S.~Yi, H.~Li, and W.~Ouyang, ``Delving into
  localization errors for monocular 3d object detection,'' in \emph{CVPR},
  2021, pp. 4721--4730.

\bibitem{messagepropagation}
L.~Wang, L.~Du, X.~Ye, Y.~Fu, G.~Guo, X.~Xue, J.~Feng, and L.~Zhang,
  ``Depth-conditioned dynamic message propagation for monocular 3d object
  detection,'' in \emph{CVPR}, 2021, pp. 454--463.

\bibitem{3dop}
X.~Chen, K.~Kundu, Y.~Zhu, A.~G. Berneshawi, H.~Ma, S.~Fidler, and R.~Urtasun,
  ``3d object proposals for accurate object class detection,'' in
  \emph{NeurIPS}, 2015, pp. 424--432.

\bibitem{3dop-pami}
X.~Chen, K.~Kundu, Y.~Zhu, H.~Ma, S.~Fidler, and R.~Urtasun, ``3d object
  proposals using stereo imagery for accurate object class detection,'' in
  \emph{IEEE T-PAMI}.\hskip 1em plus 0.5em minus 0.4em\relax IEEE, 2017, pp.
  1259--1272.

\bibitem{ocstereo}
A.~D. Pon, J.~Ku, C.~Li, and S.~L. Waslander, ``Object-centric stereo matching
  for 3d object detection,'' in \emph{ICRA}.\hskip 1em plus 0.5em minus
  0.4em\relax IEEE, 2020, pp. 8383--8389.

\bibitem{philion2020lift}
J.~Philion and S.~Fidler, ``Lift, splat, shoot: Encoding images from arbitrary
  camera rigs by implicitly unprojecting to 3d,'' in \emph{ECCV}, 2020.

\bibitem{div2020wstereo}
D.~Garg, Y.~Wang, B.~Hariharan, M.~Campbell, K.~Weinberger, and W.-L. Chao,
  ``Wasserstein distances for stereo disparity estimation,'' in \emph{NeurIPS},
  2020.

\bibitem{crossmodaldistillation}
S.~Gupta, J.~Hoffman, and J.~Malik, ``Cross modal distillation for supervision
  transfer,'' in \emph{CVPR}, 2016, pp. 2827--2836.

\bibitem{flownet}
P.~Fischer, A.~Dosovitskiy, E.~Ilg, P.~H{\"a}usser, C.~Haz{\i}rba{\c{s}},
  V.~Golkov, P.~Van~der Smagt, D.~Cremers, and T.~Brox, ``Flownet: Learning
  optical flow with convolutional networks,'' \emph{arXiv preprint
  arXiv:1504.06852}, 2015.

\bibitem{stereomagnification}
T.~Zhou, R.~Tucker, J.~Flynn, G.~Fyffe, and N.~Snavely, ``Stereo magnification:
  Learning view synthesis using multiplane images,'' \emph{arXiv preprint
  arXiv:1805.09817}, 2018.

\bibitem{bengio2011expressive}
Y.~Bengio and O.~Delalleau, ``On the expressive power of deep architectures,''
  in \emph{International conference on algorithmic learning theory}.\hskip 1em
  plus 0.5em minus 0.4em\relax Springer, 2011, pp. 18--36.

\bibitem{expressive}
J.~Kileel, M.~Trager, and J.~Bruna, ``On the expressive power of deep
  polynomial neural networks,'' \emph{Advances in Neural Information Processing
  Systems}, vol.~32, pp. 10\,310--10\,319, 2019.

\bibitem{LVIS}
A.~Gupta, P.~Dollar, and R.~Girshick, ``Lvis: A dataset for large vocabulary
  instance segmentation,'' in \emph{CVPR}, 2019, pp. 5356--5364.

\bibitem{cutpaste}
D.~Dwibedi, I.~Misra, and M.~Hebert, ``Cut, paste and learn: Surprisingly easy
  synthesis for instance detection,'' in \emph{Proceedings of the IEEE
  international conference on computer vision}, 2017, pp. 1301--1310.

\bibitem{lian2021geometry}
Q.~Lian, B.~Ye, R.~Xu, W.~Yao, and T.~Zhang, ``Geometry-aware data augmentation
  for monocular 3d object detection,'' \emph{arXiv preprint arXiv:2104.05858},
  2021.

\bibitem{zoomnet}
Z.~Xu, W.~Zhang, X.~Ye, X.~Tan, W.~Yang, S.~Wen, E.~Ding, A.~Meng, and
  L.~Huang, ``Zoomnet: Part-aware adaptive zooming neural network for 3d object
  detection,'' in \emph{AAAI}, vol.~2, 2020, p.~7.

\bibitem{disprcnn}
J.~Sun, L.~Chen, Y.~Xie, S.~Zhang, Q.~Jiang, X.~Zhou, and H.~Bao, ``Disp r-cnn:
  Stereo 3d object detection via shape prior guided instance disparity
  estimation,'' in \emph{CVPR}, 2020, pp. 10\,548--10\,557.

\bibitem{cgstereo}
C.~Li, J.~Ku, and S.~L. Waslander, ``Confidence guided stereo 3d object
  detection with split depth estimation,'' in \emph{IROS}.\hskip 1em plus 0.5em
  minus 0.4em\relax IEEE, 2020, pp. 5776--5783.

\bibitem{yolostereo3d}
Y.~Liu, L.~Wang, and M.~Liu, ``Yolostereo3d: A step back to 2d for efficient
  stereo 3d detection,'' in \emph{ICRA}.\hskip 1em plus 0.5em minus 0.4em\relax
  IEEE, 2021.

\bibitem{adam}
D.~P. Kingma and J.~Ba, ``Adam: A method for stochastic optimization,'' in
  \emph{arXiv:1412.6980}, 2014.

\bibitem{openpcdet}
O.~D. Team, ``Openpcdet: An open-source toolbox for 3d object detection from
  point clouds,'' \url{https://github.com/open-mmlab/OpenPCDet}, 2020.

\bibitem{pvrcnn}
S.~Shi, C.~Guo, L.~Jiang, Z.~Wang, J.~Shi, X.~Wang, and H.~Li, ``Pv-rcnn:
  Point-voxel feature set abstraction for 3d object detection,'' in
  \emph{CVPR}, 2020, pp. 10\,529--10\,538.

\end{thebibliography}

\end{document}